\documentclass[10pt,journal,compsoc]{IEEEtran}
\ifCLASSOPTIONcompsoc
  \usepackage[nocompress]{cite}
\else
  \usepackage{cite}
\fi
\ifCLASSINFOpdf
\else
\fi

\usepackage[utf8]{inputenc} 
\usepackage[T1]{fontenc}    
\usepackage{hyperref}       
\usepackage{url}            
\usepackage{booktabs}       
\usepackage{amsfonts}       
\usepackage{nicefrac}       
\usepackage{microtype}      
\usepackage{xcolor}         
\usepackage{microtype}
\usepackage{graphicx}
\usepackage{subfigure}
\usepackage{booktabs} 
\usepackage{multirow}
\usepackage{caption}
\usepackage{tabularx}
\usepackage[ruled,vlined,linesnumbered]{algorithm2e}
\usepackage{mathtools}
\usepackage{diagbox}
\DeclarePairedDelimiter{\norm}{\lVert}{\rVert} 
\usepackage{array}
\usepackage{amsmath}
\usepackage{amssymb}
\usepackage{amsthm}
\newtheorem{definition}{Definition}
\newtheorem{thm}{Theorem}

\newtheorem{lemma}{Lemma}
\newtheorem{cor}{Corollary}
\newtheorem{rmrk}{Remark}

\begin{document}
%
\title{Hierarchical Prototype Networks for Continual Graph Representation Learning}
%
%
%
%

\author{Xikun~Zhang, Dongjin~Song,~\IEEEmembership{Member,~IEEE}, and~Dacheng~Tao,~\IEEEmembership{Fellow,~IEEE}
\IEEEcompsocitemizethanks{
\IEEEcompsocthanksitem Xikun~Zhang  and  Dacheng~Tao  are  with the  School  of  Computer  Science, in  the  Faculty  of  Engineering,  at  The  University  of  Sydney,  Darlington,  NSW  2008,  Australia. \protect\\
Email:  xzha0505@uni.sydney.edu.au, dacheng.tao@sydney.edu.au
\IEEEcompsocthanksitem Dongjin~Song is with the Department of Computer Science and Engineering, University of Connecticut, Storrs, Connecticut, the United States.\protect\\
Email: dongjin.song@uconn.edu
}
}

\IEEEtitleabstractindextext{%
\begin{abstract}
Despite significant advances in graph representation learning, little attention has been paid to the more practical continual learning scenario in which new categories of nodes (\textit{e.g.}, new research areas in citation networks, or new types of products in co-purchasing networks) and their associated edges are continuously emerging\textcolor{black}{, causing catastrophic forgetting on previous categories.} \textcolor{black}{Existing methods either ignore the rich topological information or sacrifice plasticity for stability.} To this end, we present Hierarchical Prototype Networks (HPNs) which extract different levels of abstract knowledge in the form of prototypes to represent the continuously expanded graphs. Specifically, we first leverage a set of Atomic Feature Extractors (AFEs) to encode both the elemental attribute information and the topological structure of the target node. Next, we develop HPNs to adaptively select relevant AFEs and represent each node with three levels of prototypes. In this way, whenever a new category of nodes is given, only the relevant AFEs and prototypes at each level will be activated and refined, while others remain uninterrupted to maintain the performance over existing nodes. \textcolor{black}{Theoretically, we first demonstrate that the memory consumption of HPNs is bounded regardless of how many tasks are encountered. Then, we prove that under mild constraints, learning new tasks will not alter the prototypes matched to previous data, thereby eliminating the forgetting problem. The theoretical results are supported by experiments on five datasets, showing that HPNs not only outperform state-of-the-art baseline techniques but also consume relatively less memory.}\vspace{-1mm}
\end{abstract}

}

\maketitle

\IEEEdisplaynontitleabstractindextext

%
\IEEEpeerreviewmaketitle

\IEEEraisesectionheading{\section{Introduction}\label{sec:introduction}}

%
%
%
%
\IEEEPARstart{G}{raph} representation learning aims to pursue a meaningful vector representation of each node so as to facilitate downstream applications such as node classification, link prediction, \textit{etc}. Traditional methods are developed based on graph statistics \cite{perozzi2014deepwalk} or hand-crafted features~\cite{bhagat2011node,liben2007link}. Recently, a great amount of attention has been paid to graph neural networks (GNNs), such as graph convolutional network (GCNs)~\cite{kipf2016semi}, GraphSAGE~\cite{hamilton2017inductive}, Graph Attention Networks (GATs)~\cite{velivckovic2017graph}, and their extensions~\cite{xu2018powerful,chen2018fastgcn,zou2019layer, li2019deepgcns,chen2020simple,wang2020large}. This is because they can jointly consider the feature and topological information of each node. Most of these approaches, however, focus on static graphs and cannot generalize to the case when new categories of nodes are emerging.

In many real world applications, different categories of nodes and their associated edges (in the form of subgraphs) are often continuously emerging in existing graphs. For instance, in a citation network \cite{sen2008collective,wang2020microsoft,mikolov2013distributed}, papers describing new research areas will gradually appear in the citation graph; in a co-purchasing network such as Amazon \cite{Bhatia16}, new types of products will continuously be added to the graph. Given these facts, how to incorporate the feature and topological information of new nodes in a continuous and effective manner such that performance over existing nodes is uninterrupted is a critical problem to investigate, especially when graphs are relatively large and retraining a new model over an entire graph is computationally expensive.

\textcolor{black}{To address this issue, 
a few attempts have been made to tailor the existing continual learning techniques to graph data. 
For instance, adopting the memory~replay based approaches, Zhou et al.~\cite{zhou2021overcoming} proposed to store a set of representative experience nodes in a buffer and replay them along with new tasks (categories) to prevent forgetting existing tasks (categories). The buffer, however, only stores node features and ignores the topological information of graphs. Inspired by the regularization~based methods, Liu et al.~\cite{liu2020overcoming} developed topology-aware weight preserving~(TWP) that can preserve the topological information of existing graphs. However, its design hinders the capability of learning topology on new tasks (categories). More detailed introduction on these related works can be found in Section \ref{sec:related_work_continual_learning}.}

\textcolor{black}{A desired learning system for continual graph representation learning is expected to continuously grasp knowledge from new categories of emerging nodes and capture their topological structures without interfering with the learned knowledge over existing graphs. 
Inspired by the fact that humans learn to recognize objects by forming prototypes in the brain \cite{bowman2020tracking,dandan2013brain,zeithamova2008dissociable} and can achieve extraordinary continual learning capability,  we present a completely novel framework, \textit{i.e.}, Hierarchical Prototype Networks (HPNs), to continuously extract different levels of abstract knowledge (in the form of prototypes) from graph data such that new knowledge will be accommodated while earlier experience can still be well retained. 
Within this framework, representation learning is simultaneously conducted to avoid catastrophic forgetting, instead of considering these two objectives separately. 
Although learning independent prototypes for different tasks is appealing for preventing forgetting, it may incur unbounded memory consumption with an unlimited number of new tasks. Therefore, we propose to denote each node with a composition of basic prototypes so that nodes from different tasks and their associated relational structures are represented with compositions of a limited set of basic prototypes. Take the social network as an example, if a node (person) falls into certain category, it can be decomposed into basic atomic characteristics belonging to a set of attributes~(\textit{e.g.}, gender, nationality, hobby, \textit{etc.},) and the relationship between a pair of nodes can also be categorized into different basic types~(\textit{e.g.}, friends, coworkers, \textit{etc.}).}

\textcolor{black}{Inspired by these facts, we first develop the Atomic Feature Extractors (AFEs) to decompose each node into two sets of atomic embeddings, \textit{i.e.}, atomic node embeddings which encode the node feature information and atomic structure embeddings which encode its relations to neighboring nodes within multi-hop. Next, we present Hierarchical Prototype Networks to adaptively select, compose, and store representative embeddings with three levels of prototypes, \textit{i.e.}, atomic-level, node-level, and class-level. Given a new node, only the relevant AFEs and prototypes in each level will be activated and refined, while others are uninterrupted.} 

\textcolor{black}{The memory efficiency and the continual learning capability of the proposed HPNs are theoretically validated. First, adopting concepts from geometry and coding theory, we show the equivalence between the prototypes of HPNs and the spherical codes, thereby deriving the upper bound of the number of prototypes. This indicates that our memory consumption is bounded regardless of the number of tasks encountered. Second, since the forgetting problem can be formalized as the prediction drift of previous data after learning new tasks, we derive the conditions under which the learning on new tasks will not alter the predictions of previous data, \textit{i.e.} forgetting is eliminated.} 

\textcolor{black}{The theoretical merits of the proposed HPNs are justified by experiments on five public datasets. First, the performance of HPNs not only achieves the state-of-the-art, but is also comparable to or better than the jointly trained model (set as the upper bound of continual learning models). Second, the HPNs are memory efficient. For instance, on OGB-Products dataset containing more than 2 million nodes and 47 categories of nodes, HPNs achieve around 80$\%$ accuracy with only thousands of parameters.
} 

\section{Related Works}\label{sec:related_works}

The proposed Hierarchical Prototype Networks (HPNs) are closely related to continual learning and graph representation learning. In this section, we provide more detailed discussions by comparing HPNs with related works, especially on those directly applying existing continual learning techniques on graph data.

\subsection{Continual learning}\label{sec:related_work_continual_learning}

Continual learning aims to overcome the well-known catastrophic forgetting problem that a model's performance on previous tasks decreases significantly after being trained on new tasks. Existing works for continual learning can be categorized as
regularization-based methods, memory-replay based methods, and parametric isolation based methods.
\textcolor{black}{In the following, we give detailed introductions on these approaches, as well as their applicability to graph data.}

Regularization-based methods penalize the model objectives to maintain satisfactory performance on previous tasks \cite{jung2016less,li2017learning,kirkpatrick2017overcoming,farajtabar2020orthogonal,saha2021gradient}.
For instance, Li and Hoiem~\cite{li2017learning} introduced Learning without Forgetting (LwF) which uses knowledge distillation to constrain the shift of parameters for old tasks; Kirkpartick et al.~\cite{kirkpatrick2017overcoming} proposed elastic weight consolidation (EWC) that adds a quadratic penalty to prevent the model weights from shifting too much. Recent works \cite{farajtabar2020orthogonal,saha2021gradient} seek to constrain the gradients for new tasks in a subspace orthogonal to the updating directions that are important for previous tasks. \textcolor{black}{Although the regularization-based methods can alleviate the forgetting on previous tasks, the constraints on the model weights reduce a model's plasticity for new tasks, resulting in inferior performance compared to the other two approaches. Regularization-based methods can be directly applied to graph neural networks and are included as baselines in our experiments.}

Memory-replay based methods constantly feed a model with representative data of previous tasks to prevent forgetting \cite{lopez2017gradient,shin2017continual,aljundi2019gradient,caccia2020online,chrysakis2020online}. 
One example is Gradient Episodic Memory (GEM) \cite{lopez2017gradient} that stores representative data in episodic memory and adds a constraint to prevent the loss of the episodic memory from increasing and only allow it to decrease. Instead of storing data, Shin et al.~\cite{shin2017continual} added a generative model to generate pseudo-data of previous tasks to be interleaved with new task data for rehearsal. Recent works also look for better designs of memory replay to facilitate continual learning agents \cite{caccia2020online,chrysakis2020online}. Memory-replay based methods are currently one of the most effective approaches for alleviating catastrophic forgetting, but these methods are not suitable for graph data, as they cannot store the topological information, which is crucial for representing graph data. In contrast, our proposed HPNs explicitly designed Atomic Feature  Extractors  (AFEs) to encode both node and topological information. Moreover, memory-replayed methods require rehearsal of old data each time when a new task is learned. This introduces extra computation burdens. 

Parametric isolation based methods adaptively introduce new parameters for new tasks to avoid the parameters of previous tasks being drastically changed \cite{rusu2016progressive,yoon2017lifelong,yoon2019scalable,wortsman2020supermasks,wu2019large}.
For instance, progressive network~\cite{rusu2016progressive} allocates a new sub-network for each new task and block any modification on the previously learned networks.  Yoon et al.~\cite{yoon2017lifelong} proposed a more flexible model (DEN) that dynamically adds new neurons to accommodate new tasks. 
Recently, various innovative approaches to allocate separated parameters for different tasks have been developed \cite{wu2019large,yoon2019scalable,wortsman2020supermasks}.
Besides the concrete models, Knoblauch et al.~\cite{knoblauch2020optimal} analyzed the required capability of an optimal continual learning agent. Although parametric isolation based methods are effective for alleviating the catastrophic forgetting problem, they also consistently increase the model complexity and memory consumption which could be a problem. Recently developed methods have used certain mechanisms to control memory consumption, such as merging parameters for similar tasks \cite{yoon2019scalable}. However, as task specific parameters are fitted to each individual task, these parameters are seldom reused. In the proposed HPNs, we consider the properties of graph data and decompose each node into a combination of several prototypes. As these prototypes are largely shared by all tasks, the parameter reuse level is greatly enhanced and only limited memory needs to be consumed. Moreover, we also derive a theoretical memory upper bound for HPNs. 

Overall, although existing continual learning methods can perform well on Euclidean data, they are not suitable to be directly applied to graph data\textcolor{black}{, which is also shown in our experiments}. Our proposed HPNs are specially designed to overcome their aforementioned limitations.


\subsection{Graph representation learning}
Graph representation learning aims to encode both the feature information from nodes and the topological structure of the incoming graph. Traditional methods relied on graph statistics or hand-crafted features~\cite{bhagat2011node,liben2007link}. Recently, a great amount of attention has been paid to graph neural networks (GNNs), such as graph convolutional network (GCNs)~\cite{kipf2016semi}, GraphSAGE~\cite{hamilton2017inductive}, Graph Attention Networks (GATs)~\cite{velivckovic2017graph}, and their extensions~\cite{xu2018powerful,chen2018fastgcn,bahonar2019graph,zhang2019graph,zhang2020context,lei2020spherical,gao2021topology}. Instead of focusing on shallow networks, there are also works on building deep GCNs to further increase the capacity of GNNs~\cite{li2019deepgcns,chen2020simple,rong2019dropedge,zhang2020dropping,wang2020large}. \textcolor{black}{These models are not designed for the continual learning setting and will experience catastrophic forgetting problem when learning a sequence of tasks, which is shown in our experiments.}


Currently, only limited efforts have been made to pursuit continual graph representation. Zhou et al.~\cite{zhou2021overcoming} integrated memory-replay to GNNs by storing experience nodes from previous tasks. However, the topological structure of graphs is ignored. Liu et al.~\cite{liu2020overcoming} developed topology-aware weight preserving (TWP) that can preserve the topological information of previous graphs. However, preserving topology of previous graphs will hamper its capability of learning topology on new graphs. Galke et al.\cite{galke2020incremental} considers a scenario where new classes of nodes may appear, which is similar to our consideration. But they focus on adapting the model to new patterns and do not explicitly maintain the performance on previous tasks. Streaming GNN~\cite{wang2020streaming} and Feature graph network (FGN)~\cite{wang2020lifelong} are also related to GNNs and continual learning. However, the setting of Streaming GNN is time step incremental, while our setting is task incremental. \textcolor{black}{FGN transforms node classification into graph classification by constructing feature graphs, so as to apply existing continual learning techniques. However, the topological information is not fully utilized since the information aggregation for each node does not include the neighboring nodes.}


Note that continual graph representation learning is essentially different from dynamic graph representation learning \cite{yu2018netwalk,nguyen2018continuous,zhou2018dynamic,ma2020streaming} and few-shot graph representation learning \cite{zhou2019meta,guo2021few,yao2020graph} . Dynamic graph representation learning focuses on learning the evolving representations of nodes over time, while in a continual learning setting we constantly deal with different graphs (different tasks) and the model is not allowed to access the previously observed graphs. Different from continual learning which aims to overcome the catastrophic forgetting problem, few-shot graph representation learning targets at fast adaption to new tasks and follows a completely different setting. Specifically, for few-shot learning, a model is first trained on several meta-training tasks and then evaluated on the meta-testing tasks. During the evaluation, a few-shot learning model is independently trained and evaluated on each task, while a continual learning model learns sequentially on testing tasks and is eventually evaluated over all previous tasks. 

\section{Hierarchical Prototype Networks}

\begin{figure*}
    \centering
    \includegraphics[height=8.5cm]{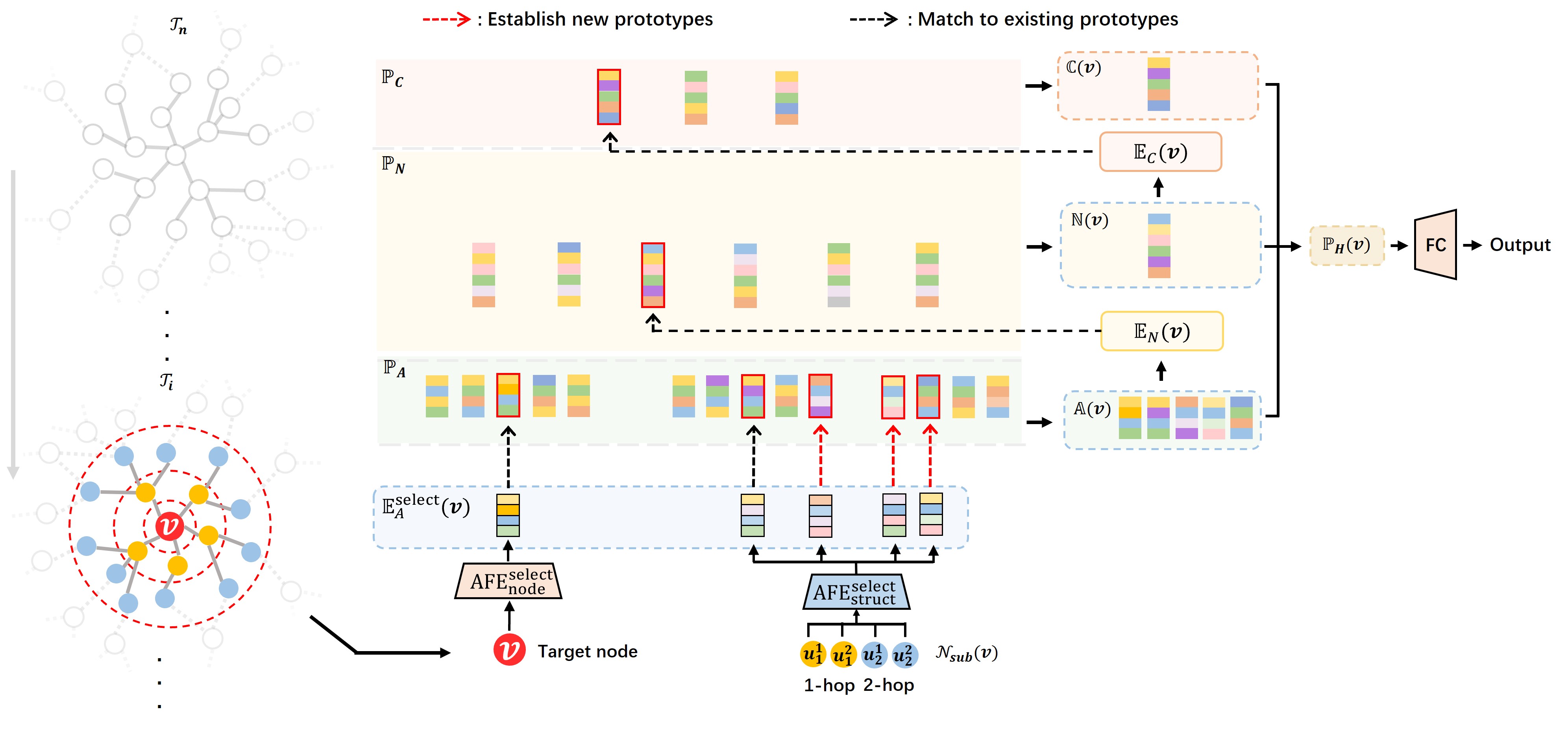}
    \captionsetup{font=small}
    \caption{The framework of HPNs. On the left, subgraphs from different tasks come in sequentially. Given a node $v$, $u_k^j$ denotes the $j$-th sampled node from $k$-hop neighbors. In the middle, node $v$ and the sampled neighbors ($\mathcal{N}_{sub}(v)$) are fed into the selected AFEs ($\mathrm{AFE}_{\textrm{node}}^{\textrm{select}}$ or $\mathrm{AFE}_{\textrm{struct}}^{\textrm{select}}$) to get atomic embeddings ($\mathbb{E}_A^{\mathrm{select}}(v)$), which are either matched to existing A-prototypes or used as new A-prototypes. The selected A-prototypes are further matched to a N- and a C-prototype for the hierarchical representation, which is finally fed into the classifier to perform node classification.}
    \label{fig:pipeline}
\end{figure*}

In this section, we first state the problem we aim to study and the notations. Then we present Hierarchical Prototype Networks (HPNs) that consist of two core modules, \textit{i.e.}, Atomic Feature Extractor (AFEs) and Hierarchical \textcolor{black}{Prototypes (HPs)}, as shown in Figure \ref{fig:pipeline}. AFEs serve to extract a set of atomic features from the given graph, and the \textcolor{black}{HPs} aim to select, compose, and store the representative features in the form of different levels of prototypes. During the training stage, each node will only refine the relevant AFEs and prototypes of the model without interfering with the irrelevant parts (\textit{i.e.}, to avoid catastrophic forgetting). In the test stage, the model will activate the relevant AFEs and prototypes to perform the inference.

\subsection{Problem Statement and Notations}
We study continual learning on graphs that have new categories of nodes and associated edges (in the form of subgraphs) emerging in a continuous manner. In the context of continual learning, assuming we have a sequence of $p$ tasks $\{\mathcal{T}^i | i=1,..., p \}$, in which each task $\mathcal{T}^i$ aims to learn a satisfied representation for a new subgraph $\mathcal{G}_i$ consisting of nodes belonging to some new categories. A desired model should maintain its performance on all previous tasks after being successively trained on the sequence of $p$ tasks from $\mathcal{T}^1$ to $\mathcal{T}^p$.

For simplicity, we omit the subscripts in this section. Full notations will be used in the theoretical analysis. Each graph $\mathcal{G}$ consists of a node set $\mathbb{V}=\{v_i | i=1,...,N\}$ with $N$ nodes and an edge set $\mathbb{E}=\{(v_i,v_j) \}$ denoting the connections of nodes in $\mathbb{V}$. Each node $v_i$ can be represented as a feature vector $\mathbf{x}(v_i)\in\mathbb{R}^{d_v}$ that encodes node attributes, \textit{e.g.}, gender, nationality, hobby, \textit{etc}. 
The set of $l$-hop neighboring nodes of $v_i$ is defined as $\mathcal{N}^l (v_i)$, with $\mathcal{N}^0 (v_i)=\{v_i\}$.

\subsection{Atomic Feature Extractors}\label{sec:AFEs}
\textcolor{black}{As mentioned in the Introduction, to handle unlimited number of potential new categories of nodes with a limited memory consumption, we represent each node as a composition of basic prototypes selected from a limited set. To construct or refine these basic prototypes, given the input graph, we first need to obtain basic features with feature extractors.}
Specifically, we develop Atomic Feature Extractors (AFEs) to consider two different sets of atomic embeddings, \textit{i.e.}, atomic node embeddings which encode the target node features and atomic structure embeddings that encode its relations to \textcolor{black}{multi-hop neighbors}. \textcolor{black}{To ensure these generated basic prototypes are capable of encoding low-level features that can be shared across different tasks, we avoid deep structures in the AFEs and formulate them} as learnable linear transformations\textcolor{black}{:} $\mathrm{AFE}_{\textrm{node}} = \{ \mathbf{A}_i \in \mathbb{R}^{d_v \times d_{a}} | i\in\{1,...,l_a\} \}$ and $\mathrm{AFE}_{\textrm{struct}} = \{ \mathbf{R}_j \in \mathbb{R}^{d_v \times d_{r}} | j\in\{1,...,l_r\} \}$ where $\mathbf{A}_i $ and $\mathbf{R}_j$ are real matrices to encode atomic node and structure information, respectively. \textcolor{black}{Each matrix $\mathbf{A}_i $ or $\mathbf{R}_j$ corresponds to a certain type of features, and $l_a$, $l_r$ denote the number of matrices in}  $\mathrm{AFE}_{\textrm{node}}$ and $\mathrm{AFE}_{\textrm{struct}}$, respectively. Given a node $v$, a set of atomic node embeddings is obtained by applying $\mathrm{AFE}_{\textrm{node}}$ to the feature vector $\mathbf{x}(v)$:
\vspace{-0cm}
\begin{align}
    \mathbb{E}^{\textrm{node}}_A(v) = \{ \mathbf{x}^T(v)\mathbf{A}_i  | \mathbf{A}_i\in\mathrm{AFE}_{\textrm{node}}\}.\label{eqn1}
\end{align}
To obtain atomic structure embeddings, the multi-hop neighboring nodes \textcolor{black}{are considered to encode the relationship between node $v$ and neighboring nodes within a fixed number of hops}. \textcolor{black}{Since the number of neighboring nodes varies from node to node, for computation efficiency in the following procedure, we uniformly sample a fixed number of vertices from 1-hop up to $h$-hop neighborhood. The sampled neighborhood set is denoted as},
$\mathcal{N}_{sub}(v) \subseteq  \bigcup\limits_{l\in\{1,...,h\}} \mathcal{N}^l (v)$. 
Then these selected \textcolor{black}{neighboring} nodes are embedded via matrices in $\mathrm{AFE}_{\textrm{struct}}$ to encode different types of interactions \textcolor{black}{between the target node $v$ and its neighbors}:
\begin{align}
     \mathbb{E}^{\textrm{struct}}_A(v) =  \{\mathbf{x}^T(u)\mathbf{R}_i | \mathbf{R}_i\in\mathrm{AFE}_{\textrm{struct}}, u\in \mathcal{N}_{sub}\}.\label{eqn2}
\end{align}
\textcolor{black}{Above all, with $\mathrm{AFE}_{\textrm{node}}$ and $\mathrm{AFE}_{\textrm{struct}}$, the attribute and structure information of each node $v$ is encoded in the two set $\mathbb{E}^{\textrm{node}}_A(v)$ and $\mathbb{E}^{\textrm{struct}}_A(v)$. For convenience of notation, we denote the complete atomic feature set of a node $v$ as the union of the node atomic embedding set $\mathrm{AFE}_{\textrm{node}}$ and the structure embedding set $\mathrm{AFE}_{\textrm{struct}}$, \textit{i.e.}},
\begin{align}
    \mathbb{E}_A(v) = \mathbb{E}^{\textrm{node}}_A(v) \cup \mathbb{E}^{\textrm{struct}}_A(v).
\end{align}
Note that \textcolor{black}{different} $\mathbf{A}_i$ and $\mathbf{R}_i$ are designed to generate different types of atomic features. \textcolor{black}{Therefore}, we impose a divergence loss on AFEs to ensure they are uncorrelated with each other and thus can map features to different subspaces:
\begin{align}
    \mathcal{L}_{div} =  \sum_{i=1}^{l_a}\sum_{\substack{j=1\\j\neq i}}^{l_a} \mathbf{A}_i^T\mathbf{A}_j + \sum_{i=1}^{l_r}\sum_{\substack{j=1\\j\neq i}}^{l_r} \mathbf{R}_i^T\mathbf{R}_j.
\end{align}

\subsection{Hierarchical Prototype Networks}\label{sec:HPs}
\textcolor{black}{Based on the atomic embeddings of the incoming subgraph, HPNs will distill and store the representative features in form of different levels of prototypes to memorize the current task}, as shown in Figure \ref{fig:pipeline}. \textcolor{black}{This is achieved by refining existing prototypes if the input subgraph only contains features observed before and creating new prototypes if new features are encountered}. Specifically, HPNs will produce three levels of prototypes. \textcolor{black}{First, the atomic-level prototypes (A-prototypes) are directly refined or created with the atomic embedding set, and they denote the prototypical basic features of nodes (\textit{e.g.}, gender, nationality, and social relation of people in a social network). Since the A-prototypes only describe low-level features of a node, which is similar to the output from the initial layers of a deep neural network, we further develop node-level (N-prototypes) and class-level prototypes (C-prototypes) to capture higher level information of each node. The N-prototypes are generated by associating the A-prototypes with each node to describe the node as a whole. The C-prototypes are generated from the N-prototypes to describe common features shared by a group of similar nodes.} From atomic-level to class-level, the prototypes denote abstract knowledge of the subgraph at different scales which is analog to the feature maps of convolutional neural networks at different layers.

\textcolor{black}{Next, we introduce when and how HPNs will refine existing prototypes or establish new ones with the atomic embeddings. As mentioned earlier, for each task that contains certain categories of nodes, only the most relevant AFEs and prototypes will be activated while others remain uninterrupted to avoid forgetting. In other words, for each node, we will choose a subset of $\mathbb{E}_A(v)$ to refine the prototypes.} Specifically, as shown in Figure \ref{fig:pipeline}, given \textcolor{black}{a node from }an incoming subgraph, \textcolor{black}{The AFEs whose atomic embeddings are close enough to existing A-prototypes are chosen as the relevant ones.} Formally, we first obtain $\mathbb{E}^{\textrm{node}}_A(v)$ and $\mathbb{E}^{\textrm{struct}}_A(v)$ via Eq. (\ref{eqn1}) and Eq. (\ref{eqn2}), respectively.
Then, we calculate the maximum cosine similarity between atomic embeddings of each AFE ($\mathbf{e}_i$) and the A-prototypes as:
\begin{align}\label{eq:simmax}
    \mathrm{SimMAX}^{\textrm{id}}_{i} = \max_{\mathbf{p}}(\frac{\mathbf{e}_i^T\mathbf{p}}{\norm{\mathbf{e}_i}_2 \norm{\mathbf{p}}_2}) , \mathbf{e}_i\in \mathbb{E}^{\textrm{id}}_A(v), \mathbf{p} \in \mathbb{P}_A,
\end{align}
where $\textrm{id} \in \{\textrm{node}, \textrm{struct}\}$, $i$ ranges from $1$ to $l_a$ (or $l_r$), \textcolor{black}{and $\mathbb{P}_A$ is the atomic prototype set containing all A-prototypes}. After that, we sort the AFEs in a descending order according to $\mathrm{SimMAX}^{\mathrm{id}}_{i}$ as $\mathrm{AFE}_{\textrm{node}}^{\textrm{sort}} = \{ \mathbf{A}_{i'} \in \mathbb{R}^{d_v \times d_{a}} | i'\in\{1,...,l_a\} \}$ and $\mathrm{AFE}_{\textrm{struct}}^{\textrm{sort}} = \{ \mathbf{R}_{j'} \in \mathbb{R}^{d_v \times d_{r}} | j'\in\{1,...,l_r\} \}$. Finally, we select the top $l_a'$ and top $l_r'$ ranked AFEs from these two sets as $\mathrm{AFE}_{\textrm{node}}^{\textrm{select}}$ and $\mathrm{AFE}_{\textrm{struct}}^{\textrm{select}}$, respectively. $l_a'$ and $l_r'$ are fixed hyperparameters with $l_a'\leq l_a$ and $l_r'\leq l_r$. \textcolor{black}{Finally, the atomic embeddings generated by these selected AFEs are denoted as $\mathbb{E}^{\textrm{select}}_A(v)$, and we have $\mathbb{E}^{\textrm{select}}_A(v) \subset \mathbb{E}_A(v)$.}
 
\textcolor{black}{The $\mathbb{E}^{\textrm{select}}_A(v)$ is then used for refining existing prototypes or establishing new ones. To determine which embeddings contain observed features and which contain new features, a} matching process is first conducted between the $\mathbb{E}^{\textrm{select}}_A(v)$ and $\mathbb{P}_A$. Formally, we measure the cosine similarity between elements in $\mathbb{E}^{\textrm{select}}_A(v)$ and elements in $\mathbb{P}_A$ as 
\begin{align}\label{eq:ea_sim}
    \mathrm{Sim}_{{E\rightarrow A}}(v) = \{\frac{\mathbf{e}_i^T\mathbf{p}}{\norm{\mathbf{e}_i}_2 \norm{\mathbf{p}}_2} | \mathbf{e}_i\in \mathbb{E}^{\textrm{select}}_A(v), \mathbf{p} \in \mathbb{P}_A \}.
\end{align}
\textcolor{black}{Given $\mathrm{Sim}_{{E\rightarrow A}}(v)$, the atomic embeddings with cosine similarity (not less than a certain threshold $t_A$) to at least one existing A-prototype are regarded as containing observed features, since they are close to existing prototypes}, \textit{i.e.},
\begin{align}
    \mathbb{E}_{old}(v) = \{\mathbf{e}_i |\quad \exists  \mathbf{p} \in \mathbb{P}_A \quad s.t.\quad \frac{\mathbf{e}_i^T\mathbf{p}}{\norm{\mathbf{e}_i}_2\norm{\mathbf{p}}_2}  \geqslant t_A \}.
\end{align}
\textcolor{black}{$\mathbb{E}_{old}(v)$ is then used to refine $\mathbb{P}_A$}. To this end, \textcolor{black}{a distance loss} $\mathcal{L}_{dis}$ is computed to enhance the cosine similarity between each $\mathbf{e}_i \in \mathbb{E}_{old}(v)$ and its corresponding A-prototype $\mathbf{p}_i \in \mathbb{P}_A$, \textit{i.e.},
\begin{align}
    \mathcal{L}_{dis} = -\sum_{\mathbf{e}_i \in \mathbb{E}_{old}(v)} \frac{\mathbf{e}_i^T\mathbf{p}_i}{\norm{\mathbf{e}_i}_2\norm{\mathbf{p}_i}_2}
\end{align}
By minimizing $\mathcal{L}_{dis}$, not only the existing A-prototypes in $\mathbb{P}_A$ will get refined, the atomic embeddings will also be closer to `standard' A-prototypes. 

Next, we discuss how to deal with the atomic embeddings that are not close to any existing prototypes, \textit{i.e.}, $\mathbb{E}_{new}(v) = \mathbb{E}^{\textrm{select}}_A(v) \backslash \mathbb{E}_{old}(v)$ or $\mathbb{E}_{new}(v) = \{\mathbf{e}_i | \quad \forall \mathbf{p} \in \mathbb{P}_A, \quad \frac{\mathbf{e}_i^T\mathbf{p}}{\norm{\mathbf{e}_i}_2\norm{\mathbf{p}}_2} < t_A \}$.

Contrary to $\mathbb{E}_{old}(v)$, atomic embeddings in $\mathbb{E}_{new}(v)$ are regarded as new atomic features of the corresponding AFEs \textcolor{black}{ and should be accommodated with new prototypes.} \textcolor{black}{Considering that very similar embeddings may exist in $\mathbb{E}_{new}(v)$ and cause HPNs to create redundant prototypes, we first filter $\mathbb{E}_{new}(v)$ to obtain $\mathbb{E}'_{new}(v)$ which only contains the distinctive ones such that}
\begin{align}
    \forall \mathbf{e}_i, \mathbf{e}_j \in \mathbb{E}'_{new}(v), \frac{\mathbf{e}_i^T\mathbf{e}_j}{\norm{\mathbf{e}_i}_2\norm{\mathbf{e}_j}_2}  < t_A.
\end{align}
In this way, \textcolor{black}{the embeddings in $\mathbb{E}_{new}'(v)$ only contain new features, they are included into $\mathbb{P}_A$ as new A-prototypes. Formally, the updated $\mathbb{P}_A$ is the union of the previous A-prototypes $\mathbb{P}_A$ and new A-prototypes $\mathbb{E}_{new}'(v)$}
\begin{align}\label{eq:pa_update}
    \mathbb{P}_A = \mathbb{P}_A \cup \mathbb{E}'_{new}(v).
\end{align}
After \textcolor{black}{updating $\mathbb{P}_A$, $\mathrm{Sim}_{{E\rightarrow A}}(v)$ will also be updated according to Eq. (\ref{eq:ea_sim}). Then, every element in $\mathrm{Sim}_{{E\rightarrow A}}(v)$ would be not less than $t_A$, \textit{i.e.} each element in $\mathbb{E}^{\textrm{select}}_A(v)$ is matched to an A-prototype with cosine similarity larger than the threshold $t_A$. These matched A-prototypes are denoted as $\mathbb{A}(v)$, which represents the basic features of node $v$ as shown in Figure \ref{fig:pipeline}.}

\textcolor{black}{As mentioned earlier, besides $\mathbb{A}(v)$, higher level representations are also beneficial for comprehensively describing a node from multiple granularities. Therefore, we further map $\mathbb{A}(v)$ to high level prototypes so as to obtain hierarchical prototype representations,} Firstly, $\mathbb{A}(v)$ is mapped to an N-prototype \textcolor{black}{to describe $v$ as a whole. Specifically, the vectors in $\mathbb{A}(v)$ are concatenated as $\mathbf{a}_{1}\oplus\cdots\oplus\mathbf{a}_{l_a'+l_r'}$ ($\mathbf{a}_i \in \mathbb{A}(v)$), and then fed into a full connected layer $\textrm{FC}_{A\rightarrow N}(\cdot)$ to generate node-level embedding set $\mathbb{E}_N(v) = \{\textrm{FC}_{A\rightarrow N}(\mathbf{a}_{1}\oplus\cdots\oplus\mathbf{a}_{l_a'+l_r'}) \}$. Here, for notation consistency with the aforementioned atomic embedding set, we also formulate the node-level embeddings as a set $\mathbb{E}_N(v)$ although it contains only one element. With $\mathbb{E}_N(v)$, the following procedure is to form a N-prototype representation $\mathbb{N}(v)$ for node $v$, which is same as the process to form $\mathbb{A}(v)$ at atomic level except that the threshold is set as $t_N$, instead of $t_A$. Finally, to enrich the node representation with the common features shared with similar nodes, we develop the C-prototypes. The generation of C-prototypes is similar to the N-prototypes. We first obtain a class-level embedding set $\mathbb{E}_C(v)$ by applying another fully connected layer $\textrm{FC}_{N\rightarrow C}$ to the element in $\mathbb{N}(v)$. Then, we match $\mathbb{E}_C(v)$ to existing C-prototypes or establish new ones. Different from N-prototypes, the threshold used here is $t_C$, which is typically larger than $t_A$ and $t_N$. In this way, each C-prototype will be matched to embeddings within a large space, and therefore representing the common features shared by a wide range of similar nodes. 
Again, for notation convenience, we derive $\mathbb{P}_H(v)$ to represent the hierarchical prototype representations of the target node $v$ (as shown in Figure \ref{fig:pipeline}), which is the union of $\mathbb{A}(v)$, $\mathbb{N}(v)$, and $\mathbb{C}(v)$: }
\begin{align}\label{eq:ph}
    \mathbb{P}_H(v) = \mathbb{A}(v) \cup \mathbb{N}(v) \cup \mathbb{C}(v).
\end{align}

\begin{algorithm}[t]
\small
\captionsetup{font=small}
\caption{Learning Procedure for HPNs.}
\label{tqnn_qnn_classification_algorithm}
\SetKwInOut{Input}{Input}
\SetKwInOut{Output}{Output}
\Input{Task sequence: $\{\mathcal{T}_1,...,\mathcal{T}_p\}$, HPNs}
\For{$\mathcal{T}\gets1$ \KwTo $p$}{
    Get the data of the current task: $\mathbb{V}$, $\mathbb{E}$, $\mathbf{X}(\mathbb{V})=\{\mathbf{x}(v)|v \in \mathbb{V}\}$.
    
    Select $\mathrm{AFE}_{\textrm{node}}^{\textrm{select}}$ and $\mathrm{AFE}_{\textrm{struct}}^{\textrm{select}}$.
    
    Compute $\mathcal{L}$ = HPNs($\mathbb{V}$,$\mathbf{X}(\mathbb{V})$,$\mathbb{E}$). 
    
    Optimize $\mathcal{L}$.
    }
\Output{updated HPNs}
\end{algorithm}
\subsection{Learning Objective}
The obtained hierarchical prototypes of each node are first concatenated into a unified vector and then forward through a fully connected layer $\mathrm{FC}$ to obtain a $c$ (the number of classes) dimensional feature vector, \textit{i.e.}, $\textrm{FC}(\mathbf{h}_{1}\oplus\cdots\oplus\mathbf{h}_{l_a'+l_r'+2}), \forall \mathbf{h}_i\in\mathbb{P}_H(v)$. In this paper, we aim to perform node classification. Therefore, based on the $c$ dimensional feature vector and the softmax function $\sigma(\cdot)$, we can estimate the label with $\hat{y}_i=\sigma(\textrm{FC}(\mathbf{h}_{1}\oplus\cdots\oplus\mathbf{h}_{l_a'+l_r'+2}))_i$ where $i$ is the index of class. To perform node classification, with the output predictions $\hat{y}_i$ and the target label $y_i\in \{1,2,...,c\}$, the corresponding classification loss is given by
\begin{align}
    \mathcal{L}_{cls} =\sum_{i=1}^c -y_i\log(\hat{y_i}),
\end{align}
which is essentially the cross entropy loss function. Note that besides node classification, $\mathbb{P}_H(v)$ may also be used for other tasks based on different objective functions. In this paper, we focus on node classification and the overall loss of HPNs is:
\begin{align}
\label{eq:loss}
    \mathcal{L} =\mathcal{L}_{cls} + \alpha \cdot \mathcal{L}_{div} + \beta \cdot \mathcal{L}_{dis},
\end{align}
where $\alpha\geq 0$ and $\beta\geq 0$ are hyperparameters to balance the two auxiliary losses.

\subsection{Theoretical Analysis}\label{sec:theoretical_analysis}
In this subsection, we provide the theoretical upper bound on the memory consumption and analyze how the model configuration would affect HPNs' capacity in dealing with different tasks. Both theoretical results are justified and analyzed in the experiments. Only the main results are provided here, while the detailed proof and analysis are given in supplementary materials.

We first formulate the upper bound on the numbers of prototypes.
Due to the mechanism to create new prototypes for newly emerging knowledge extracted from the graph, the memory consumption will gradually increase. However, the normalization applied to the prototypes constrains the prototype space, and thereby adds an upper bound on the memory consumption. This can be intuitively understood as the only limited number of points can be scattered on a n-dimensional hypersphere if we force the distance between each pair of points to be larger than a threshold. To formally formulate this, we first borrow the definition of spherical code from geometry and coding theory. 

\begin{definition}[Spherical code]\label{def:spherical code}
A spherical code $S(n,N,t)$, with parameters $(n,N,t)$ is defined as the set of $N$ points on the unit hypersphere in an $n$-dimension space for which the dot product of unit vectors from the origin to any two points is larger than or equal to $t$.
\end{definition}

In HPNs, the prototypes at different levels are normalized into unit vectors and can be viewed as spherical codes in their own space. Taking the A-prototypes as an example, given the dimension $d_a$ and the threshold $t_A$, $\mathbb{P}_A$ is a spherical code $\mathbb{P}_A = S_A(d_a, N_A, 1 - t_A)$, where $N_A$ is the cardinality of $\mathbb{P}_A$. Then the upper bound on the number of atom prototypes given $d_a$ and $t_A$ is equal to the maximal cardinality of $S(n,N,t)$ given $n$ and $t$. Since the area of the $n$-dimensional unit sphere surface is limited, there exists a maximal $N$ given a certain $n$ and $t$, denoted as $ \max_{N} S(d_a, N, 1-t_A)$. However, finding $ \max_{N} S(d_a, N, 1-t_A)$ is a complex sphere packing problem, and there is not yet a general formulation of the maximal $N$ for an arbitrary $n$. Therefore, given the number of two types of AFEs as $l_a$ and $l_r$, we can formulate the upper bound on the numbers of different prototypes as:

\begin{thm}[Upper bounds on numbers of prototypes]\label{thm:upper_bound}
Given the notations defined in HPNs, the upper bound on the number of A-prototypes $n_a$ can be given by
\begin{align}
    n_A \leqslant (l_a + l_r) \max_{N} S(d_a, N, 1-t_A),
\end{align}
and the upper bounds on the number of N-prototypes and the C-prototypes are:
\begin{align}
    n_N \leqslant \max_{N} S(d_n, N, 1-t_N) \\ n_C \leqslant \max_{N} S(d_c, N, 1-t_C) 
\end{align}
where $S(n,N,t)$ is the spherical code defined on a $n$ dimensional hypersphere .
\end{thm}

Theorem \ref{thm:upper_bound} provides an upper bound on the memory consumption of HPNs. In our experiments, we show that the number of parameters for most baseline methods are even higher than this upper bound.

Besides memory consumption, the more important problem for a continual learning model is the capability to maintain memory on previously learned tasks. Based on our model design, we formulate this as: whether learning new tasks affect the representations the model generates for old task data. We give explicit definitions on tasks and task distances based on set theory (in supplementary materials), then construct a bound to indicate what configuration would the model have to ensure this capability.

\begin{thm}[Task distance preserving]\label{thm:task distance preserve}
For $\mathrm{HPNs}$ trained on consecutive tasks $\mathcal{T}^p$ and $\mathcal{T}^{p+1}$. If $l_ad_a+l_rd_r \geqslant (l_r+1)d_v $ and $\mathbf{W}$ is column full rank, then as long as
$ t_{A} < \lambda_{\min}(l_r+1) \mathbf{dist}(\mathbb{V}_p, \mathbb{V}_{p+1}) $,  learning on $\mathcal{T}^{p+1}$ will not modify representations $\mathrm{HPNs}$ generate for data from $\mathcal{T}^p$
, i.e., catastrophic forgetting is avoided. 
\end{thm}

The proof of Theorem \ref{thm:task distance preserve} requires a number of definitions, lemmas, and corollaries, therefore the details are provided in the supplementary materials. The theoretical results also provide insights into model implementations. In Theorem \ref{thm:task distance preserve}, $\lambda_i$ is eigenvalues of the $\mathbf{W}^T \mathbf{W}$, where $\mathbf{W}$ denotes the matrix constructed via AFEs (details in supplementary materials). $d_v$, $d_a$ and $d_r$ are dimensions of data and two kinds of atomic embeddings. The bound in this theorem is not tight, since the tight bound would be dependent on the specific dataset properties. But this informs us that either the number of AFEs or the dimension of the prototypes has to be large enough to ensure that data from two tasks can be well separated in the representation space. 
Then, according to Theorem \ref{thm:upper_bound}, the upper bound of the memory consumption is dependent on $S(d_a,N, t_A)$, $S(d_n,N, t_N)$, and $S(d_c,N, t_C)$. As $S(n,N,t)$ grows fast with $n$, we prefer larger number of $\mathrm{AFE}$s with smaller prototype dimensions. We also empirically demonstrate this in Section \ref{sec:sensitivity}. Besides, the upper bound proposed in Theorem \ref{thm:upper_bound} is explicitly computed and compared to experimental results.

\section{Experiments}
In the experiments, we answer the following six questions: (1) Whether HPNs can outperform state-of-the-art approaches? (2) How does each component of HPNs contribute to its performance? (3) Whether HPNs can memorize previous tasks after learning each new task?
(4) Are HPNs sensitive to the hyperparameters? (5) Whether the theoretical results can be empirically verified? (6) Whether the learned prototypes can be interpreted via visualization?
All experiments are repeated on 5 different datasets. Due to space limit, only main results are presented in this paper, while the details over some datasets are included in the supplementary materials.
\begin{table}[h]
\scriptsize
    \centering
        \caption{The detailed statistics of 5 datasets used in our experiments.}
    \begin{tabular}{lccccc}\toprule
        \textbf{Dataset} & Cora & Citeseer  & Actor & OGB-Arxiv & OGB-Products\\\midrule
        \# nodes         & 2,708 & 3,327      & 7,600  & 169,343    & 2,449,029 \\\midrule
        \# edges      & 5,429 & 4,732      & 33,544 &1,166,243 &61,859,140 \\ \midrule
        \# features    & 1,433 & 3,703      & 931 & 128 & 100 \\\midrule
        \# classes         & 7    & 6         & 4 & 40 & 47 \\\midrule
        \# tasks          & 3    & 3         & 2 & 20 & 23 \\\bottomrule
    \end{tabular}
    \label{tab:data_statistics}
\end{table}

\subsection{Datasets}
To assess the effectiveness of the proposed HPNs, we consider 5 public datasets which include 3 citation networks (Cora \cite{sen2008collective}, Citeseer\cite{sen2008collective}, OGB-Arxiv \cite{wang2020microsoft,mikolov2013distributed}), 1 actor co-occurrence network (Actor) \cite{pei2020geom}, and 1 product co-purchasing networks (OGB-Products \cite{Bhatia16}). The statistics of the 5 datasets are summarized in Table \ref{tab:data_statistics}. Key properties of the datasets are presented here, and more details are included in the supplementary materials.

\subsubsection{Citation networks}
Cora contains 2708 documents, 5429 links denoting the citations among the documents. For training, 140 documents are selected with 20 examples for each class. The validation set contains 500 documents and the test set contains 1000 examples. In our continual learning setting, the first 6 classes are selected and grouped into 3 tasks (2 classes per task) in the original order.
Citeseer contains 3312 documents and 4732 links. 20 documents per class are selected for training, 500 documents are selected for validation, and 1000 documents are selected as the test set. For continual learning setting, the documents from 6 classes are grouped into 3 tasks with 2 classes per task in the original order.
The Cora and Citeseer datasets can be downloaded via \href{https://github.com/tkipf/gcn/tree/master/gcn/data}{Cora$\&$Citeseer}.

The OGB-Arxiv dataset is collected in the Open Graph Benchmark \href{https://ogb.stanford.edu/docs/nodeprop/#ogbn-arxiv}{OGB}. 
It is a directed citation network between all Computer Science (CS) arXiv papers indexed by MAG \cite{wang2020microsoft}. 
Totally it contains 169,343 nodes and 1,166,243 edges. The dataset contains 40 classes. As the dataset is imbalanced and the numbers of examples in different classes differs significantly, directly grouping the classes into 2-class groups like the Cora and Citeseer will cause the tasks to be imbalanced. Therefore, we reordered the classes in an descending order according to the number of examples contained in each class, and then group the classes according to the new order. In this way, the number of examples contained in different classes of each task are arranged to be as balanced as possible. Specifically, the class indices of each task are: \{(35, 12), (15, 21), (28, 30), (16, 24), (10, 34), (8, 4), (5, 2), (27, 26), (36, 19), (23, 31), (9, 37), (13, 3), (20, 39), (22, 6), (38, 33), (25, 11), (18, 1), (14, 7), (0, 17), (29, 32)\}, where each tuple denotes a task consisting of two classes.

\subsubsection{Actor co-occurrence network}
The actor co-occurrence network is a subgraph of the film-director-actor-writer network \cite{tang2009social}. 
Each node in this dataset corresponds to an author, and the edges between the nodes are co-occurrence on the same Wikipedia pages. 
The whole dataset contains 7600 nodes and 33544 edges. Each node is accompanied with a feature vector of 931 dimensions. For this dataset, we also constructed 2 tasks with 2 classes per task.
The link to this dataset is \href{https://github.com/graphdml-uiuc-jlu/geom-gcn/tree/master/new_data/film}{Actor}. 
The balanced splitting of the classes is \{(0, 1), (2, 3)\}.

\begin{table*}[]
    \centering
    \captionsetup{font=small}
    \caption{Performance comparisons between HPNs and baselines on 5 different datasets.}
    \scriptsize 
    \begin{tabular}{c|c||cc|cc|cc|cc|cc}
    \toprule
    \multirow{2}{2.5em}{\textbf{C.L.T.}} & \multirow{2}{2em}{\textbf{Base}}& \multicolumn{2}{c|}{Cora} &  \multicolumn{2}{c|}{Citeseer} & \multicolumn{2}{c|}{Actor} &  \multicolumn{2}{c|}{OGB-Arxiv} &  \multicolumn{2}{c}{OGB-Products} \\ \cline{3-12}
                                &     & AM/\% & FM/\% & AM/\% & FM/\% & AM/\%       & FM /\%       & AM/\%       & FM /\%       & AM/\%       & FM /\% \\ \bottomrule\toprule
     \multirow{3}{2.5em}{None}  & GCN &63.5$\pm$1.9 &-42.3$\pm$0.4 &64.5$\pm$3.9 & -7.7$\pm$1.6 & 43.6$\pm$3.6 & -9.1$\pm$2.9 & 56.8$\pm$4.3  & -19.8$\pm$3.2  & 45.2$\pm$5.6 & -27.8$\pm$7.1   \\
                                & GAT &71.9$\pm$3.8 &-33.1$\pm$2.3 &66.8$\pm$0.9 &-19.6$\pm$0.3 & 53.1$\pm$2.7 & -4.3$\pm$1.6 & 54.3$\pm$3.5 & -21.7$\pm$4.6& 44.9$\pm$6.9 & -30.3$\pm$5.2 \\  
                                & GIN &68.3$\pm$2.3 &-35.4$\pm$3.4 &57.7$\pm$2.3 &-36.4$\pm$0.3 & 45.5$\pm$2.3 & -8.9$\pm$2.8 & 53.2$\pm$6.5 & -23.5$\pm$8.1 & 43.1$\pm$7.4 & -31.4$\pm$8.8  \\  \midrule
     \multirow{3}{2.5em}{EWC \cite{kirkpatrick2017overcoming}}   & GCN &63.1$\pm$1.2 &-42.7$\pm$1.6 &54.4$\pm$4.2 &-30.3$\pm$0.9 & 44.3$\pm$1.1 & -7.1$\pm$1.4 & 72.1$\pm$2.4 & -9.1$\pm$1.9   & 66.7$\pm$0.5 & -8.4$\pm$0.4\\  
                                & GAT &72.2$\pm$1.5 &-32.2$\pm$1.6 &65.7$\pm$2.5 &-19.7$\pm$2.3 & 54.2$\pm$2.5 &-2.5$\pm$1.5 &73.2$\pm$1.1 &-10.8$\pm$2.1  & 67.9$\pm$1.0 & -9.6$\pm$1.3 \\  
                                & GIN &69.6$\pm$2.6 &-28.5$\pm$2.8 &57.9$\pm$3.4 &-36.3$\pm$2.4 & 47.6$\pm$2.1 &-7.2$\pm$1.6 &74.1$\pm$1.7 &-8.3$\pm$2.0  & 67.3$\pm$2.3 & -13.6$\pm$1.5 \\  \midrule
     \multirow{3}{2.5em}{LwF \cite{li2017learning}}   & GCN &76.1$\pm$1.4 &-21.3$\pm$2.4 &67.0$\pm$0.2 &-8.3$\pm$2.7 & 49.7$\pm$2.5 &-3.6$\pm$1.4 &69.9$\pm$3.9& -12.1$\pm$2.8 & 66.3$\pm$2.5 & -11.8$\pm$3.4 \\  
                                & GAT &70.8$\pm$2.8 &-34.6$\pm$4.1 &66.1$\pm$4.1 &-18.9$\pm$1.5 & 52.8$\pm$2.7 &-6.2$\pm$2.2 & 68.9$\pm$4.4 & -13.6$\pm$3.3 & 65.1$\pm$4.1 & -13.2$\pm$2.9  \\
                                & GIN &74.1$\pm$2.7 &-23.3$\pm$0.8 &63.1$\pm$1.9 &-16.5$\pm$2.2 &49.7$\pm$2.6 &-4.1$\pm$1.5 &71.4$\pm$4.8 & -15.9$\pm$5.6 & 65.9$\pm$4.0 & -10.7$\pm$3.1 \\ \midrule
     \multirow{3}{2.5em}{GEM \cite{lopez2017gradient}}   & GCN &75.7$\pm$3.0 &-6.5$\pm$4.4  &41.8$\pm$2.6 &-31.9$\pm$1.4 &52.7$\pm$3.1 & +3.9$\pm$2.9 & 75.4$\pm$1.7 & -13.6$\pm$0.5 & 71.3$\pm$1.7 & -10.5$\pm$0.9  \\  
                                & GAT &69.8$\pm$3.0 &-26.1$\pm$2.6 &71.3$\pm$2.2 &+9.0$\pm$1.5 &54.3$\pm$3.6 &-2.0$\pm$0.9 &76.6$\pm$0.7 & -11.3$\pm$0.4 & 70.4$\pm$0.8 & -10.9$\pm$1.6  \\  
                                & GIN &80.2$\pm$3.3 &-2.0$\pm$4.2  &49.7$\pm$0.5 &-24.5$\pm$0.9 &45.2$\pm$2.8&-11.1$\pm$1.5 &77.3 $\pm$2.1 & -11.2$\pm$1.6 & 76.5$\pm$3.3 & -7.2$\pm$2.5 \\  \midrule
     \multirow{3}{2.5em}{MAS \cite{aljundi2018memory}}   & GCN &65.5$\pm$1.9 &-21.4$\pm$3.7 &59.5$\pm$3.1 &-0.1$\pm$2.4 &50.7$\pm$2.4&-1.5$\pm$0.8 &69.8$\pm$0.4 & -18.8$\pm$0.9 & 62.0$\pm$1.1 & -17.9$\pm$1.9\\  
                                & GAT &84.7$\pm$0.7 &-5.6$\pm$2.0  &69.1$\pm$1.1 &-4.8$\pm$3.3 &53.7$\pm$2.6&-1.6$\pm$0.8 &70.6$\pm$1.3 &-16.7$\pm$1.6 & 64.4$\pm$2.3 & -14.5$\pm$3.2\\
                                & GIN &76.7$\pm$2.6 &-4.0$\pm$3.6  &65.2$\pm$3.9 &+0.0$\pm$2.0 &51.6$\pm$2.6&-0.6$\pm$1.3 &65.3 $\pm$2.9 & -17.0$\pm$2.3 & 61.4$\pm$3.8 & -20.9$\pm$2.9\\ \midrule
     \multirow{3}{2.5em}{ERGN. \cite{zhou2020continual}} & GCN &63.5$\pm$2.4 &-42.3$\pm$0.7 &54.2$\pm$3.9 &-30.3$\pm$1.9 &52.4$\pm$3.3&+0.6$\pm$1.4 & 63.3$\pm$1.7 & -18.1$\pm$0.9 & 60.7$\pm$2.8 & -26.6$\pm$3.3 \\ 
                                & GAT &71.1$\pm$2.5 &-34.3$\pm$1.0 &65.5$\pm$0.3 &-20.4$\pm$3.9 &51.4$\pm$2.2&-7.2$\pm$3.2 & 63.5$\pm$2.4 & -19.5$\pm$1.9 & 61.3$\pm$1.7 & -25.1$\pm$0.8 \\ 
                                & GIN &68.3$\pm$0.4 &-35.4$\pm$0.4 &57.7$\pm$3.1 &-36.4$\pm$1.3 &42.7$\pm$3.9&-13.0$\pm$2.1& 69.2$\pm$1.8& -11.8$\pm$1.4 & 61.8$\pm$4.7 & -23.4$\pm$7.9\\ \midrule
     \multirow{3}{2.5em}{TWP \cite{liu2020overcoming}}   & GCN &68.9$\pm$0.9 &-5.7$\pm$1.5  &60.5$\pm$3.8 &-0.3$\pm$4.4 &50.6$\pm$2.0&-4.8$\pm$1.1 & 75.6$\pm$0.3 & -10.4$\pm$0.5 & 69.9$\pm$0.4 & -9.0$\pm$1.1\\ 
                                & GAT &81.3$\pm$3.2 &-14.4$\pm$1.5 &69.8$\pm$1.5 &-8.9$\pm$2.6 &54.0$\pm$1.8&-2.1$\pm$1.9 & 75.8$\pm$0.5 & -5.9$\pm$0.3  & 69.3$\pm$2.3 & -8.9$\pm$1.5 \\ 
                                & GIN &73.7$\pm$3.2 &-3.9 $\pm$2.6 &68.9$\pm$0.7 &-2.4$\pm$1.9 &49.9$\pm$1.9&-3.6$\pm$2.0 & 76.6$\pm$1.8 & -11.3$\pm$1.1 & 69.9$\pm$1.4 & -10.3$\pm$2.7\\ \midrule
     \multirow{3}{2.5em}{Join.} & GCN &93.7$\pm$ 0.5 & - & 78.9$\pm$0.4  & -   &57.0$\pm$0.9  & -  &77.2$\pm$0.8 & - &72.9$\pm$1.2 &- \\ 
                                & GAT &93.9$\pm$ 0.9 & - & 79.3$\pm$0.8 & -  &57.1$\pm$0.9  & -  & 81.8$\pm$0.3 & -& 73.7$\pm$2.4 & -\\ 
                                & GIN &93.2$\pm$ 1.2 & - & 78.7$\pm$0.9& -   &56.9$\pm$0.6 & - &82.3$\pm$1.9 &- &77.9$\pm$2.1 &- \\
\bottomrule\toprule
     \multicolumn{2}{c||}{\textbf{HPNs}}&\textbf{93.7$\pm$1.5} &\textbf{+0.6$\pm$1.0} &\textbf{79.0$\pm$0.9} &\textbf{-0.6$\pm$0.7} &\textbf{56.8$\pm$1.4} & \textbf{ -0.9$\pm$0.9} & \textbf{85.8$\pm$ 0.7} & \textbf{+0.6$\pm$0.9}& \textbf{80.1$\pm$0.8}  &\textbf{+2.9$\pm$1.0}  \\ \bottomrule
    \end{tabular}
    \label{tab:comparison}
\end{table*}

\subsubsection{Product co-purchasing network}
OGB-Products is collected in the Open Graph Benchmark \href{https://ogb.stanford.edu/docs/nodeprop/#ogbn-arxiv}{OGB}, representing an Amazon product co-purchasing network \href{http://manikvarma.org/downloads/XC/XMLRepository.html}{link}. It contains 2,449,029 nodes and 61,859,140 edges. Nodes represent products sold in Amazon, and edges indicate that the connected products are purchased together. In our experiments, we select 46 classes and omit the last class containing only 1 example. Similar to OGB-Arxiv, we reorder the classes and the class indices of each tasks are: \{(4, 7), (6, 3), (12, 2), (0, 8), (1, 13), (16, 21), (9, 10), (18, 24), (17, 5), (11, 42), (15, 20), (19, 23), (14, 25), (28, 29), (43, 22), (36, 44), (26, 37), (32, 31), (30, 27), (34, 38), (41, 35), (39, 33), (45, 40)\}.


\subsection{Experimental Setup and Evaluation Metrics}
To perform continual graph representation learning with new categories of nodes continuously emerging, like the traditional continual learning works, we split each dataset into multiple tasks and train the model on them sequentially. Each new task brings a subgraph with new categories of nodes and associated edges, \textit{e.g.}, task 1 contains classes 1 and 2, task 2 contains new classes 3 and 4, \textit{etc.} Each model is trained on a sequence of tasks, and the performance will be evaluated on all previous tasks. Specifically, we adopt accuracy mean (AM) and forgetting mean (FM) as metrics for evaluation. After learning on all tasks, the AM and FM are computed as the average accuracy and the average accuracy decrease on all previous tasks. Negative FM indicates the existence of forgetting
, zero FM denotes no forgetting and positive FM denotes positive knowledge transfer between tasks.
For HPNs, we set $d_a = d_n=d_c=16$, $l_a=l_r=22$, and $h=2$. The threshold $t_A$, $t_N$, and $t_C$ are selected by cross validation on $\{0.01, 0.05,0.1,0.15,0.2,0.25,0.3,0.35, 0.4\}$. The experiments on the important hyperparameters are provided in Section~\ref{sec:sensitivity}. \textcolor{black}{All experiments are repeated 5 times with Nvidia Titan Xp GPU. The average performance and standard deviations are reported. More details and code are provided in supplementary materials.}

\begin{figure*}
    \centering
    \includegraphics[height = 4.0cm]{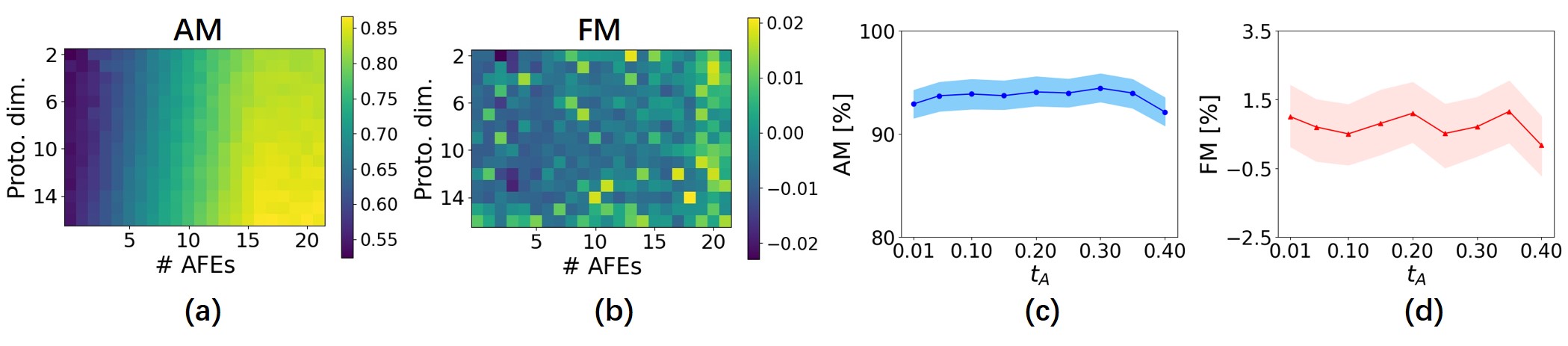}
    \captionsetup{font=small}
    \caption{(a) and (b) are AM and FM of HPNs with different number of $\mathrm{AFE}$s and prototype dimensions on OGB-Arxiv. (c) and (d) are AM and FM change with when $t_A$ varies on Cora.}
    \label{fig:AM_mat}
\end{figure*}

\begin{table}[]
\scriptsize
    \centering
    \begin{tabular}{c|ccc||ccc}\toprule
         & \multicolumn{3}{c||}{\textbf{GAT}}&\multicolumn{3}{c}{\textbf{HPNs}} \\\midrule
       \backslashbox{Te.}{Tr.}&$\mathcal{T}_1$ & $\mathcal{T}_{1,2}$&$\mathcal{T}_{1,2,3}$&$\mathcal{T}_1$&$\mathcal{T}_{1,2}$&$\mathcal{T}_{1,2,3}$ \\\midrule
       $\mathcal{T}_1$ &94.12\% &47.96\% &71.49\% &95.02\% &95.93\% &94.57\% \\\midrule
       $\mathcal{T}_2$ &      &93.30\% &49.68\% &&92.66\% &92.22\% \\\midrule
       $\mathcal{T}_3$ &  & &94.44\% &&&96.83\% \\\bottomrule
    \end{tabular}
    \caption{Accuracy (\%) changes of GAT and HPNs.}
    \label{tab:individual_tasks_GAT_DHPN}
\end{table}
\begin{table}[]
\scriptsize
    \centering
    \begin{tabular}{c|ccc||ccc}\toprule
         & \multicolumn{3}{c||}{\textbf{GCN+GEM}}&\multicolumn{3}{c}{\textbf{GAT+MAS}} \\\midrule
       \backslashbox{Te.}{Tr.}&$\mathcal{T}_1$ & $\mathcal{T}_{1,2}$&$\mathcal{T}_{1,2,3}$&$\mathcal{T}_1$&$\mathcal{T}_{1,2}$&$\mathcal{T}_{1,2,3}$ \\\midrule
       $\mathcal{T}_1$ &93.67\% &87.33\% &76.92\% &94.12\% &90.50\% &90.05\% \\\midrule
       $\mathcal{T}_2$ &      &65.66\% &69.33\% &&77.97\% &70.84\% \\\midrule
       $\mathcal{T}_3$ &  & &80.95\% &&&93.25\% \\\bottomrule
    \end{tabular}
    \caption{Accuracy (\%) changes of GCN+GEM and GAT+MAS.}
    \label{tab:individual_tasks_GCNGEM_GATMAS}
\end{table}

\subsection{Comparisons with Baseline Methods}\label{sec:comparison}
We compare HPNs with various baseline methods. Experience Replay based GNN~(ERGNN) \cite{zhou2021overcoming} and Topology-aware Weight Preserving~(TWP) \cite{liu2020overcoming} are developed for continual graph representation learning. The others approaches, including Elastic Weight Consolidation (EWC) \cite{kirkpatrick2017overcoming}, Learning without Forgetting (LwF) \cite{li2017learning}, Gradient Episodic Memory (GEM) \cite{lopez2017gradient}, and Memory Aware Synapses (MAS) \cite{aljundi2018memory}) are popular continual learning methods for Euclidean data. All the baselines are implemented based on three popular backbone models, \textit{i.e.}, Graph Convolutional Networks (GCNs) \cite{kipf2016semi}, Graph Attentional Networks (GATs) \cite{velivckovic2017graph}, and  Graph Isomorphism Network (GIN) \cite{xu2018powerful}. 

Note that Joint training (Join.) in Table~\ref{tab:comparison} does not represent continual learning. It allows a model to access data of all tasks at any time and thus is often used as an upper bound for continual learning.\cite{van2019three}. Therefore, FM is not applicable to Joint training approach.

In Table \ref{tab:comparison}, we observe that regularization based approaches, \textit{e.g.}, EWC and TWP, generally obtain lower forgetting, but the accuracy (AM) is limited by the constraints. However, the forgetting problem of regularization based methods will become increasingly severe when the number of tasks is relatively large, as shown in Section~\ref{sec:learning_dynamics}.
Memory replay based methods such as GEM achieve better performance without using any constraint. However, the memory consumption is higher (Section~\ref{sec:memory_consumption}). 
HPNs significantly outperform all baselines without inheriting their limitations. Compared to regularization based methods, HPNs do not impose constraints to limit the model's expressiveness, therefore the performance is much better. Compared to memory replay based methods, HPNs do not only perform better but also are memory efficient as shown in Section \ref{sec:memory_consumption}. 
Joint training (Join.) achieves comparable performance to HPNs on small datasets but is significantly worse on large OGB datasets. This is because joint training (Join.) is a multi-task setting, inter-task interference may cause negative transfer, which is not obvious on small datasets with only a few tasks but becomes prominent on large datasets with tens of tasks. In HPNs, different tasks can choose different combinations of the parameters and thus task interference is dramatically alleviated.

\begin{table}[]
\scriptsize 
    \centering
    \captionsetup{font=small}
    \caption{Ablation study on prototypes of different levels of prototypes over Cora.}
    \begin{tabular}{c|c|c|c|c|c}
    \toprule
     Conf. &A-p.     & N-p.    & C-p.    & AM\%           & FM\% \\ \midrule
     1  &\checkmark &           &           & 89.2$\pm$1.3   & -0.1$\pm$0.5 \\ \midrule
     2  &\checkmark &\checkmark &           & 91.7$\pm$1.1   & -0.2$\pm$0.8  \\ \midrule
     3  &\checkmark &\checkmark &\checkmark & 93.7$\pm$1.5   & +0.6$\pm$1.0 \\ \bottomrule
    \end{tabular}
    \label{tab:ablation_proto}
\end{table}
\begin{table}[]
\scriptsize 
    \centering
    
    \captionsetup{font=small}
    \caption{Ablation study on different loss terms over Cora.}
    \begin{tabular}{c|c|c|c|c|c}
    \toprule
     Conf.  &$\mathcal{L}_{cls}$ & $\mathcal{L}_{div}$ & $\mathcal{L}_{dis}$ & AM\%        & FM\%        \\ \midrule
     1 &\checkmark           &                     &                     &92.4$\pm$1.3 &+0.8$\pm$0.7  \\ \midrule
     2 &\checkmark           & \checkmark          &                     &92.9$\pm$1.1 &+0.3$\pm$1.0 \\ \midrule
     3 &\checkmark           &                     &\checkmark           &92.8$\pm$0.9 &+0.0$\pm$1.2  \\ \midrule
     4    &\checkmark           &\checkmark           &\checkmark           &93.7$\pm$1.5 &+0.6$\pm$1.0  \\ \bottomrule
    \end{tabular}
    \label{tab:ablation_loss}
    \label{tab:ablation}
\end{table}

Among all the models in Table \ref{tab:comparison}, we could observe several kinds of forgetting behavior among the baselines. First, some of them are with low AM and severe forgetting (negative FM with large $|\textrm{FM}|$) like the pure GNNs in the first 3 rows of Table 1 in the paper. These models may perform well on individual tasks, but the AM is brought down by catastrophic forgetting. To explain this in details, we expand the results of GAT without any continual learning techniques in Table \ref{tab:individual_tasks_GAT_DHPN}, in which Tr. denotes on which tasks has the model been trained and Te. denotes on which task the model is tested. 

The first row of Table \ref{tab:individual_tasks_GAT_DHPN} shows the model performance on task 1 after it has been trained on the task 1, on the first two tasks, and on the first three tasks. 
We see that GAT performs well on each individual task it has just learnt. But after being trained on more tasks, the performance on previous tasks drops dramatically, making the AM to be relatively low. 
In contrast, the performance change on previous tasks of HPNs is also shown in Table \ref{tab:individual_tasks_GAT_DHPN}, and we could see that HPNs can well maintain the performance on previous tasks throughout the training on all tasks.

Second, there are also some models without severe forgetting but the still low AM. Typically these are the models which preserve the performance on previous tasks with certain constraints on the models. These constraints can indeed alleviates forgetting, but it also limits the model's flexibility in learning new tasks, and thus the performance on new tasks will degrade. For example, comparing Table \ref{tab:individual_tasks_GAT_DHPN} with Table \ref{tab:individual_tasks_GCNGEM_GATMAS}, GAT achieves accuracy higher than 93$\%$ on individual tasks, but for GAT+MAS, although the forgetting on task 1 is alleviated, the performance on task 2 drops significantly. GCN+GEM also suffers from the similar problem.

\begin{figure*}[t]
    \centering
    \begin{minipage}{0.33\textwidth}
        \centering
        \includegraphics[width=0.9\textwidth]{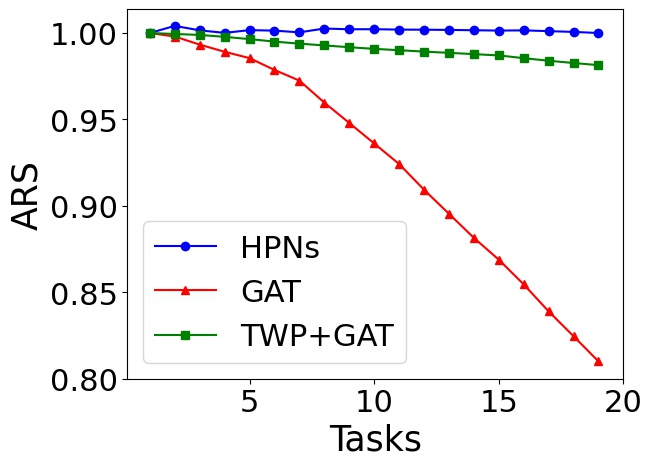}
    \end{minipage}\hfill
    \begin{minipage}{0.33\textwidth}
        \centering
        \includegraphics[width=0.9\textwidth]{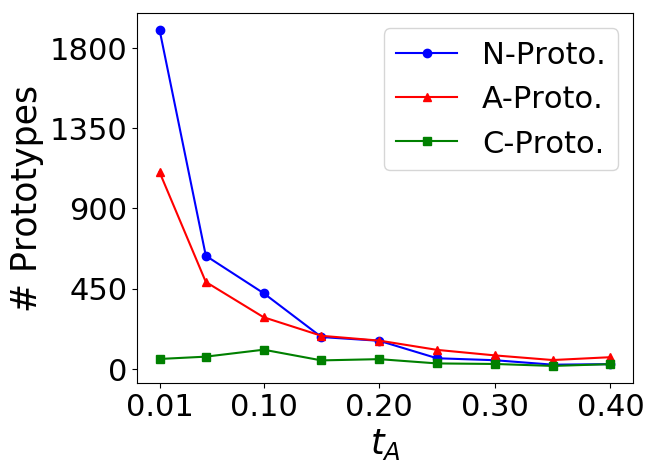}
    \end{minipage}
    \begin{minipage}{0.33\textwidth}
        \centering
        \includegraphics[width=0.9\textwidth]{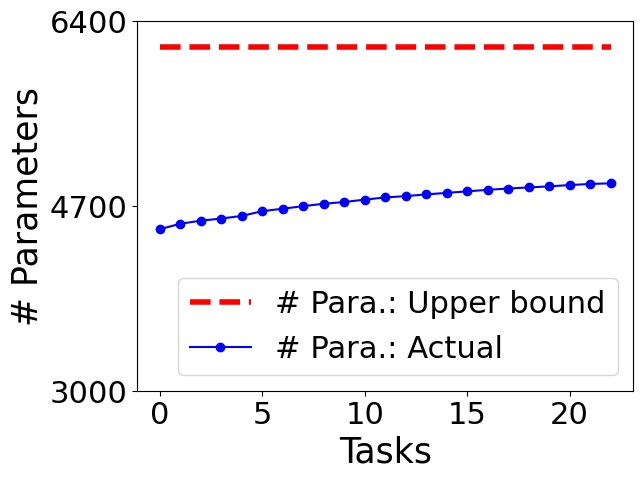}
    \end{minipage}
    \captionsetup{font=small}
     \caption{Left: dynamics of ARS for continual learning tasks on OGB-Arxiv. Middle: impact of $t_A$ on the number of prototypes in HPNs over Cora. Right: dynamics of memory consumption of HPNs on OGB-Products.}
     \label{fig:ARS_para_proto}
\end{figure*}

\begin{table*}[!t]
\scriptsize 
    \centering
    \captionsetup{font=small}
    \caption{ Final parameter amount for models trained on OGB-Products}
    \begin{tabular}{c|c|c|c|c|c|c|c|c||c}
    \toprule
            & None  & EWC    & LwF   & GEM    & MAS  & ERGNN & TWP  & Joint & \textbf{HPNs} \\ \midrule
      GCN   &  2,336 & 46,720  & 4,672  &2,202,336 &2,336  &6,738   &9,344  & 2,336  & \\ 
      GAT   & 20,032 & 400,640 & 40,064 &2,220,032 &20,032 &24,432  &80,128 & 20,032 &4,908\\ 
      GIN   &  2,352 & 47,040  & 4,704  &2,202,352 & 2,352 &6,752   &9,408  & 2,352  & \\ \midrule
    \end{tabular}
    \label{tab:memory_consumption}
\end{table*}

\subsection{Ablation Study}
We conduct ablation studies on different levels of prototypes and different combinations of three loss terms. In Table \ref{tab:ablation_proto}, we show the performance of HPNs when A-, N-, and C-Prototypes are gradually added (Cora dataset). 
We notice both AM and FM of HPNs increase when higher level prototypes are considered. This suggests that high level prototypes can enhance the model's performance and robustness against forgetting.
The effect of different combinations of loss terms are shown in Table \ref{tab:ablation_loss}. The first three rows show that adding $\mathcal{L}_{div}$ or $\mathcal{L}_{dis}$ with $\mathcal{L}_{cls}$ may slightly improve the performance. By jointly considering these three terms, the performance (AM) can be further improved. This is because $\mathcal{L}_{div}$ pushes different $\mathrm{AFE}$s away from each other and $\mathcal{L}_{dis}$ makes the prototypes of each $\mathrm{AFE}$ be more close to its output. Jointly considering $\mathcal{L}_{div}$ and $\mathcal{L}_{dis}$ with $\mathcal{L}_{cls}$ can make the prototype space better separated as shown in Section~\ref{sec:visualization}.

\subsection{Learning Dynamics}\label{sec:learning_dynamics}

For continual learning, it is important to memorize previous tasks after learning each new task. To measure this, instead of directly measuring the average accuracy on previous tasks which may mix up the accuracy change caused by forgetting and task differences, we develop a new metric, \textit{i.e.}, average retaining score (ARS), to address this problem. Specifically, after learning on a task $\mathcal{T}^i$, the ratio between the model's accuracy on a previous task $\mathcal{T}^{i-m}$ and its accuracy on $\mathcal{T}^{i-m}$ after it had been just learned on $\mathcal{T}^{i-m}$ is defined as the retaining ratio. Then the ARS is the average retaining ratio of all previous tasks after learning a new task. 

\begin{figure*}[]
    \centering
    \begin{minipage}{0.245\textwidth}
        \centering
        \includegraphics[width=1.\textwidth]{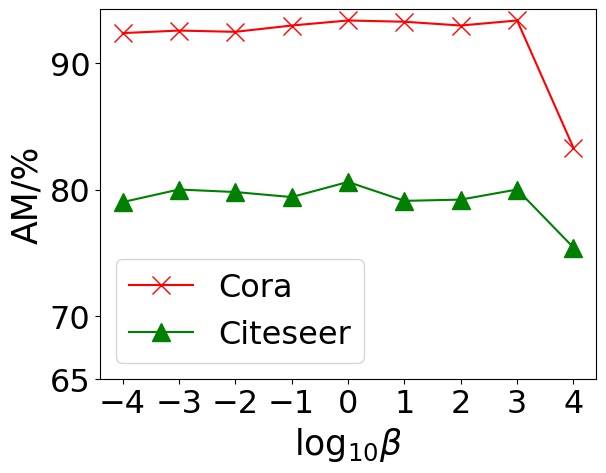}
    \end{minipage}
    \begin{minipage}{0.245\textwidth}
        \centering
        \includegraphics[width=1.\textwidth]{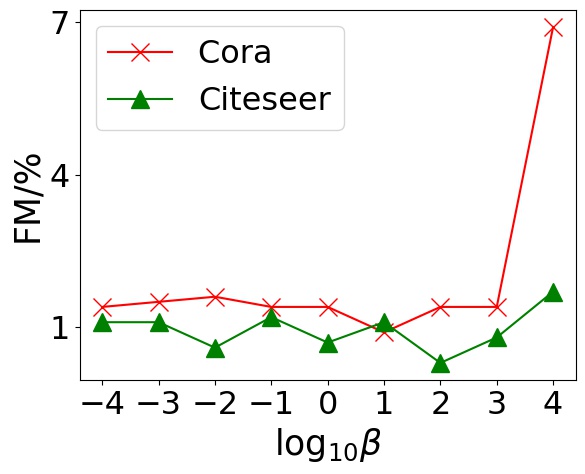}
    \end{minipage}
    \begin{minipage}{0.245\textwidth}
        \centering
        \includegraphics[width=1.\textwidth]{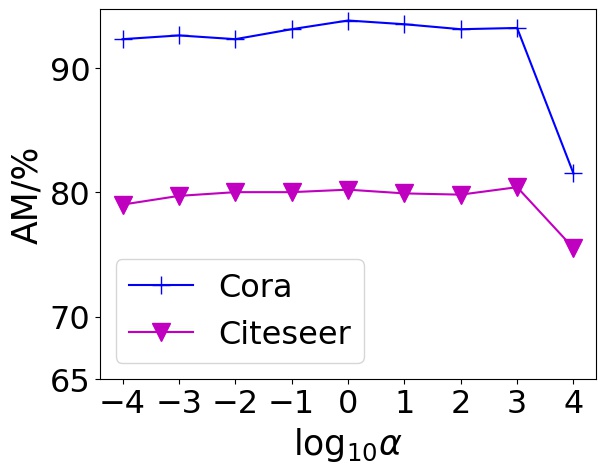}
    \end{minipage}
    \begin{minipage}{0.245\textwidth}
        \centering
        \includegraphics[width=1.\textwidth]{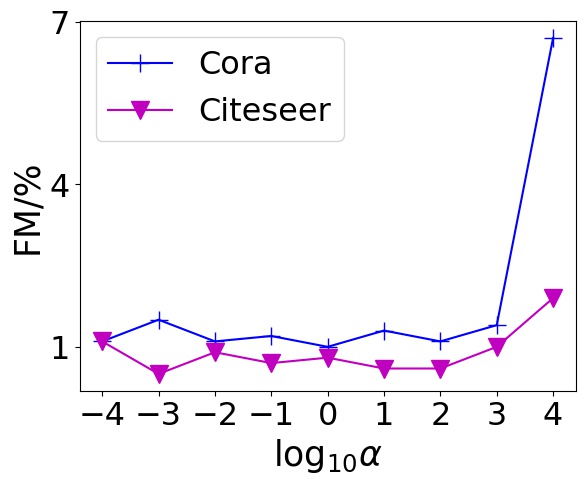}
    \end{minipage}
    \captionsetup{font=small}
     \caption{The left two figures show the results of fixing $\alpha=1.0$ and tuning $\beta$ from 0.0001 to 1000. The right two figures show the results of fixing $\beta=1.0$ and tuning $\alpha$. Logarithmic horizontal axis is adopted, and the results are obtained on both Cora and Citeseer datasets.}
     \label{fig:loss_coef}
\end{figure*}

Figure \ref{fig:ARS_para_proto}(left) shows the ARS change of HPNs and two baselines. GAT represents the models without continual learning techniques. TWP+GAT is the best baseline in terms of forgetting. GAT forgets quickly, while TWP significantly alleviates the forgetting problem for GAT. But as more tasks come in, the forgetting of TWP+GAT increases. As different tasks require different parameters, TWP+GAT (regularization based) is seeking a trade-off between old and new tasks. With more new tasks, TWP+GAT tends to gradually adapt to new tasks and forget old ones. On contrary, HPNs maintain the ARS very well. This is because HPNs learn prototypes to denote the common basic features and learning new tasks does not hurt the parameters for old tasks. New tasks can be handled with new combinations of the existing basic prototypes. If necessary, new prototypes can be established for more expressiveness.

\subsection{Parameter Sensitivity}\label{sec:sensitivity}
As discussed in Section \ref{sec:theoretical_analysis}, the number of $\mathrm{AFE}$s and the prototype dimensions are key factors in determining the continual learning capability and memory consumption. Here, we conduct experiments with different number of $\mathrm{AFE}$s and prototype dimensions to justify the theoretical results. We keep the dimensions of different prototypes equal and the number of two types of $\mathrm{AFE}$s equal for simplicity. 

As shown in Figure \ref{fig:AM_mat}(a) and (b), larger dimensions and the number of $\mathrm{AFE}$s yield better AM and FM, which is consistent with Theorem \ref{thm:task distance preserve}. Besides, AM is mostly determined by the number of $\mathrm{AFE}$s since HPNs compose prototypes with different $\mathrm{AFE}$s to represent each target node. The number of possible combinations determines its expressiveness. 
Considering the above results and the bound (Theorem \ref{thm:upper_bound}) for the number of prototypes, using large number of $\mathrm{AFE}$s and small dimension can ensure both high performance and low memory usage, as verified in Section \ref{sec:memory_consumption}.

We also evaluate the effectiveness of HPNs when prototype thresholds vary from 0.01 to 0.4. In Figure \ref{fig:AM_mat}(c) and (d), we observe that the performance (AM and FM) of HPNs are generally stable when $t_A$ varies and slightly better when $t_A$ is between 0.2 and 0.3. This is because when $t_A$ is too small or too large, we will have too many or too less prototypes (consistent with Theorem \ref{thm:upper_bound}) as shown in Figure \ref{fig:ARS_para_proto}(middle), which may cause the problem of overfitting or underfitting.

\begin{figure*}[h]
    \centering
    \begin{minipage}{0.31\textwidth}
        \centering
        \includegraphics[width=1.\textwidth]{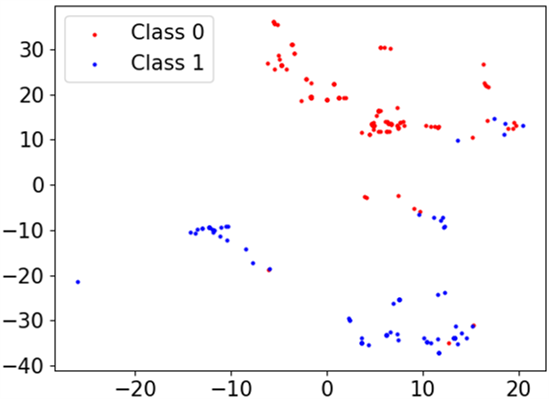}
    \end{minipage}
    \begin{minipage}{0.31\textwidth}
        \centering
        \includegraphics[width=1.\textwidth]{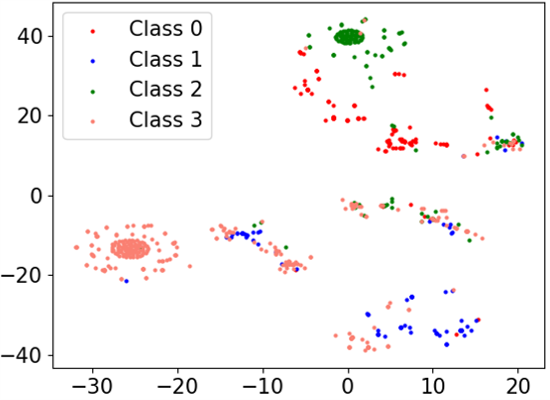}
    \end{minipage}
    \begin{minipage}{0.31\textwidth}
        \centering
        \includegraphics[width=1.\textwidth]{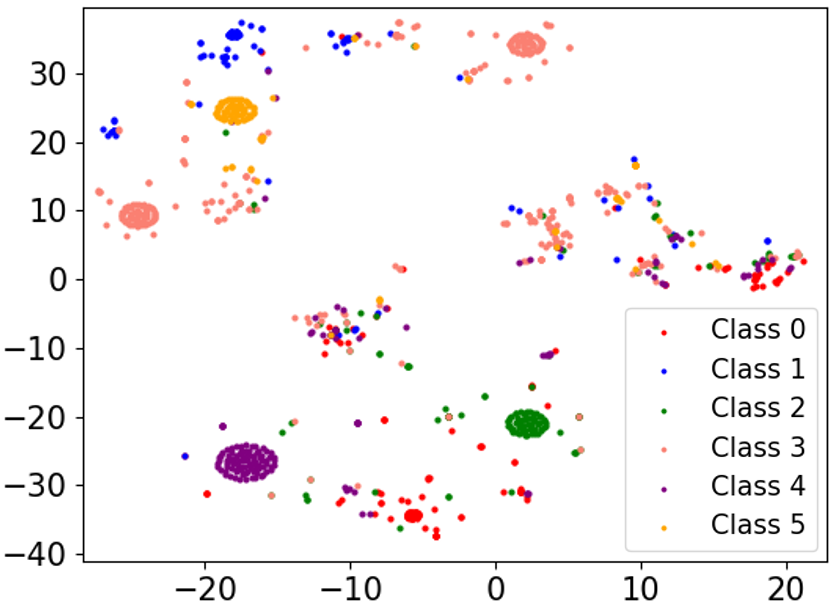}
    \end{minipage}
    \captionsetup{font=small}
     \caption{Visualization of hierarchical prototype representations of nodes in the test set of Cora.}
     \label{fig:tsne}
\end{figure*}

Finally, we explore the influence of the hyperparameters of the auxiliary losses ($\alpha$ and $\beta$ in Equation \ref{eq:loss}) on the performance. As shown in Figure \ref{fig:loss_coef}, we fix $\alpha=1.0$ or $\beta=1.0$ in turn, and tune the other one to check the parameter sensitivity. 

From Figure \ref{fig:loss_coef}, we can observe that the performance is relatively robust when $\alpha$ and $\beta$ vary from 0.0001 to 1000.
Through monitoring the training process, we suggest the reason is that $\mathcal{L}\_{div}$ and $\mathcal{L}\_{dis}$ decrease faster than $\mathcal{L}\_{cls}$, thus the total loss is not very sensitive to the coefficients on them. Based on this observation, we choose to simply set $\alpha=\beta=1$ in implementations.

In this subsection, we systematically studied the dependency of the model on the hyperparameters. According to the results, HPNs are robust against most of the hyperparameters. The number of AFEs will cause significant difference but the relation between the performance and the number of AFEs is simple. For a given dataset, the performance first increases with the number of AFEs, and becomes stable when the number AFEs reaches a certain threshold (\textit{i.e.} large enough).

\subsection{Memory Consumption}\label{sec:memory_consumption}
We compare memory consumption of different methods, as well as a explicitly theoretical memory upper bound, with the baselines on OGB-Products (the largest dataset). We also show the actual memory consumption of HPNs in the process of continual learning.

In Table \ref{tab:memory_consumption}, even on the dataset with millions of nodes and 23 tasks, HPNs can accommodate all tasks with a small amount of parameters. Besides, the dynamic change of parameter amount is shown in Figure \ref{fig:ARS_para_proto}(right). The red dashed line denotes the theoretical upper bound (6,163), and the computation details are included in supplemental materials.
In Figure \ref{fig:ARS_para_proto}(right), we notice the actual memory usage of HPNs is much lower than the upper bound. Moreover, even the upper bound is among the lowest for memory consumption compared to baselines. The model we use here is the same as the one in Section \ref{sec:comparison}

\subsection{Visualization}\label{sec:visualization}
To concretely show the prototype representations generated by HPNs, we apply t-SNE \cite{van2008visualizing} to visualize the node representations of the Cora dataset (test set) after learning each task, as shown in Figure \ref{fig:tsne}. The three figures display the representations generated after each of the three tasks arrives sequentially. Each task contains two classes denoted with different colors, \textit{i.e.} (red, blue), (green, salmon), and (purple, orange). In Figure \ref{fig:tsne}, with new tasks coming in sequentially from left to right, the representations are consistently well separated, which is beneficial for downstream tasks.

\section{Conclusion}

In this paper, we proposed Hierarchical Prototype Networks (HPNs), to continuously extract different levels of abstract knowledge (in the form of prototypes) from streams of tasks on graph representation learning. The continual learning performance of HPNs is both theoretically analyzed and experimentally justified from different perspectives on multiple public datasets with up to millions of nodes and tens of millions of edges. Besides, the memory consumption of HPNs is also theoretically analyzed and experimentally verified. 

HPNs provide a general framework for continual graph representation learning and can be further extended to tackle more challenging tasks. In the future, we will extend HPNs to heterogeneous graphs by adapting the AFEs and prototypes to capture multiple types of nodes and edges. Moreover, we will also equip the prototypes with connections to denote the relationship among the abstract knowledge and investigate more challenging graph reasoning tasks.

\section{\textcolor{black}{Proof Details}}\label{sec:theoretical_analysis}
\subsection{Overview}
The main theoretical results are briefly introduced in the paper. In this section, we provide detailed explanations and proofs for the theoretical results. 

\subsection{Memory consumption upper bound}
The proof and detailed analysis on the memory consumption upper bound has already been given in the paper. In this subsection, we give the computation of a specific case of the general theoretical results.

Although the general formulation of the upper bound is not available, we can specially compute $\max_{N} S(d_a, N, 1-t_A)$ for certain $n$s, and verify it with experiments. For example, when $n=2$, the distribution becomes distributing points on a circle with unit radius. Then, $\max_{N} S(d_a, N, 1-t_A)$ can be obtained by evenly distributing the points on the circle with an interval of $t_A$. Finally, the explicit value of $\max_{N} S(d_a, N, 1-t_A)$ can be formulated as:

\begin{align}
    \max_{N} S(d_n, N, 1-t_N) = \frac{2\pi}{\arccos(1-t_A)},
\end{align}
then we have:
\begin{align}
    n_A \leqslant (l_a + l_r) \frac{2\pi}{\arccos(1-t_A)}.
\end{align}

And the upper bound of the number of N- and C-prototypes can be formulated similarly. The above results are used in Section 4.7 in the paper. 

\subsection{Task distance preserving}
In this subsection, we give proofs and detailed analysis on the Theorem 2 in the paper.
At below, Lemma \ref{quadrtic_bound}, Lemma \ref{real_symmetric}, Lemma \ref{rank}, and Corollary \ref{cor:1} are from existing knowledge ranging from geometry to linear algebra. The other parts are of our own contributions.

In continual learning, the key challenge is to overcome the catastrophic forgetting, which refers to the performance degradation on previous tasks after training the model on new tasks. Based on our model design, we formulate this as: whether learning new tasks affect the representations the model generates for old task data. First, we give definitions on the tasks and task distances:

\begin{definition}[Task set]\label{def:task_set}
The $p$-th task in a sequence is denoted as $\mathcal{T}^p$ and contains a subgraph $\mathcal{G}_p$ consisting of nodes belonging to some new categories. We denote the associated node set and adjacency matrix as $\mathbb{V}_p$ and $\mathrm{A}_p$. 
Each $v_p^i \in \mathbb{V}_p$ has a feature vector $\mathbf{x}(v_p^i)$ and a label $y(v_p^i)$.
\end{definition}
Then, the reason for catastrophic forgetting is that different tasks in a sequence are drawn from heterogeneous distributions, making the model sequentially trained on different tasks 
unable to maintain satisfying performances on previous tasks. Therefore, given the definition of the tasks (Definition \ref{def:task_set}), we then give a formal definition to quantify the difference between two tasks.

\begin{definition}[Task distance]\label{def:task_distance}
We define the distance between two tasks as the set distance between the node sets of these two tasks, \textit{i.e.}

$\mathbf{dist}(\mathbb{V}_p, \mathbb{V}_q) = \mathrm{inf}\norm{\mathbf{x}(v_p^i) - \mathbf{x}(v_q^j)}, \forall v_p^i \in \mathbb{V}_p, v_q^j \in \mathbb{V}_q$.
\end{definition}

\begin{lemma}
The distance between any two tasks is non-negative, i.e. $\forall i,j\in \{1,...,\mathrm{M}^T\}, \mathbf{dist}(\mathbb{V}_p, \mathbb{V}_q) \geqslant 0$, where $\mathrm{M}^T$ is the number of tasks contained in the sequence.
\end{lemma}

The real-world data could be complex and sometimes may even contain noises that are impossible for any model to learn, which needs extra considerations when justifying the effectiveness of the model. Formally, we give the definition of the contradictory data.

\begin{definition}[Contradictory data]
$\forall v_p^i \in \mathbb{V}_p, p = 1,...,\mathrm{M}^T$, if $\exists v_q^j \in \mathbb{V}_q, j = 1,...,\mathrm{M}^T$, $\mathrm{st.} \forall l\in \mathbb{N^*}, \forall u \in \mathcal{N}^l(v_p^i) $ and $\forall v \in \mathcal{N}^l(v_q^j)$, $\mathbf{x}(u)=\mathbf{x}(v)$ but $y(v_p^i) \neq y(v_q^j)$, then we say $(v_p^i, y(v_p^i))$ and $(v_q^j, y(v_q^j))$ are contradictory data, as it is contradictory for any model to give different predictions for one node based on the same node features and graph structures. ($\mathbb{N}^*$ denotes the set of non-negative integers)
\end{definition}

\begin{rmrk}
Contradictory data is ignored or simply regarded as outliers in previous works, but in this work, we explicitly analyze its affect for the comprehensiveness of our theory.
contradictory data has different situations. 
If $v_p^i$ and $v_q^j$ are from different tasks, then $y(v_p^i) \neq y(v_q^j)$ is plausible. Because they may be describing a same thing from different aspects. For example, an article from the citation network may be both categorized as 'physics related' and 'computer science related'. In this situation, it would be easy to add an task indicator to the feature of the node, then the feature of $v_p^i$ and $v_q^j$ are no longer equal and are not contradictory data anymore. 

But within one task, contradictory data are most likely to be wrongly labeled, \textit{e.g.} it does not make sense if an article is both 'related to physics' and 'not related to physics'. 
\end{rmrk}

Besides the distance between tasks, the distance between the embeddings obtained by the AFEs will also be a crucial concept in the proof.

\begin{definition}[Embedding distance]
Each input node $v_p^i$ is given a set of atomic embeddings $\mathbb{E}_A(v_p^i) = \mathbb{E}^{\mathrm{node}}_A(v_p^i) \cup \mathbb{E}^{\mathrm{struct}}_A(v_p^i)$, where $\mathbb{E}^{\mathrm{node}}_A(v_p^i) = \{\mathbf{a}_i^j | j\in\{1,...,l_a\}\}_p $ containing the atomic node embeddings of $v_p^i$ and $\mathbb{E}^{\mathrm{struct}}_A(v_p^i) = \{\mathbf{r}_i^j | k \in \{ 1,...,l_r\} \}_p$ containing the atomic structure embeddings. $\mathbf{a}_i^j\in \mathbb{R}^{d_a}$ and $\mathbf{r}_i^j \in \mathbb{R}^{d_r}$. To define the distance between representations of two nodes, we concatenate the atomic embeddings of each node into a single vector in a higher dimensional space, i.e. each node $v_p^i$ corresponds to a latent vector $\mathbf{z}_p^i=[\mathbf{a}_i^1;...;\mathbf{a}_i^{l_a};\mathbf{r}_i^1;...;\mathbf{r}_i^{l_r}]\in\mathbb{R}^{l_a \times d_a + l_r \times d_r}$. Then we define the distance between representations of two nodes $v_p^i$ and $v_q^j$ as the Euclidean distance between their corresponding latent vector $\mathbf{z_p^i}$ and $\mathbf{z_q^j}$, i.e. $\mathrm{dist}(\mathbf{z}_p^i,\mathbf{z}_q^j) = \norm{\mathbf{z}_p^i-\mathbf{z}_q^j}_2$ .
\end{definition}

Then we will give some explanations on the linear algebra related theories.

\begin{lemma}[Bounds for real quadratic forms]\label{quadrtic_bound}
Given a real symmetric matrix $\mathbf{A}$, and an arbitrary real vector variable $\mathbf{x}$, we have
\begin{align}
    \lambda_{\min} \leqslant \frac{\mathbf{x}^T \mathbf{A} \mathbf{x}}{\mathbf{x}^T \mathbf{x}} \leqslant \lambda_{\max},
\end{align}
where $\lambda_{\min}$ and $\lambda_{\max}$ are the minimum and maximum eigenvalues of matrix $\mathbf{A}$.
\end{lemma}

\begin{lemma}[Real symmetric matrix]\label{real_symmetric}
For a matrix $\mathbf{A} \in \mathbb{R}^{m\times n}$, $\mathbf{A}^T\mathbf{A} \in \mathbb{R}^{n\times n}$ is a real symmetric matrix, $rank(\mathbf{A}^T\mathbf{A}) = rank(\mathbf{A})$, and the non-zero eigenvalues of $\mathbf{A}^T\mathbf{A}$ are squares of the non-zero singular values of $\mathbf{A}$.
\end{lemma}

\begin{lemma}[Rank and number of non-zero singular values]\label{rank}
For a matrix $\mathbf{A} \in \mathbb{R}^{m\times n}$, the number of non-zero singular values equals the rank of $\mathbf{A}$, i.e. $rank(\mathbf{A})$
\end{lemma}

\begin{cor}\label{cor:1}
For a matrix $\mathbf{A} \in \mathbb{R}^{m\times n}$. Without loss of generality, we assume $n \leqslant m$. If $\mathbf{A}$ is column full rank, i.e. $rank(\mathbf{A})=n$, then $\mathbf{A}$ has $n$ non-zero singular values. Besides, $rank(\mathbf{A}^T\mathbf{A}) = n$, and $\mathbf{A}^T\mathbf{A}$ has $n$ non-zero singular values.
\end{cor}

Given the explanations above, we then derive the bound for the change of the distance among data, which will be further used for analyzing the separation of data from different tasks.

\begin{lemma}[Embedding distance bound]\label{dist_bound}
Given two nodes $v_p^i \in \mathbb{V}_p$ and $v_q^j \in \mathbb{V}_q$ with vertex feature $\mathbf{x}(v_p^i), \mathbf{x}(v_q^j) \in \mathbb{R}^{{d_v}}$, their multi-hop neighboring node sets are denoted as $\bigcup\limits_{l\in \mathbb{N}^*} \mathcal{N}^l (v_p^i)$ and $\bigcup\limits_{l\in \mathbb{N}^*} \mathcal{N}^l (v_q^j)$. The AFEs for generating atomic embeddings are $\mathrm{AFE}_{\mathrm{node}} = \{ \mathbf{A}_i \in \mathbb{R}^{d_{a} \times d_v } | i\in\{1,...,l_a\} $ and $\mathrm{AFE}_{\mathrm{struct}} = \{ \mathbf{R}_j \in \mathbb{R}^{d_r \times d_v } | j\in\{1,...,l_r\}$, corresponding to matrices for atomic node embeddings and atomic structure embeddings, respectively. Then, the square distance 

$\mathrm{dist}^2(\mathbf{z}_p^i - \mathbf{z}_q^j) = \norm{\mathbf{z}_p^i - \mathbf{z}_q^j}_2^2 \geqslant \lambda_{\min}(\norm{\mathbf{x}(v_p^i)-\mathbf{x}(v_q^j)}_2^2 + \sum_{k=1}^{l_r}\norm{\mathbf{x}(u_k)-\mathbf{x}(\nu_k)}_2^2)$,

if $l_a \times d_a + l_r \times d_r \geqslant d_v $, where $u_k$ are nodes sampled from $\bigcup\limits_{l\in \mathbb{N}^*} \mathcal{N}^l (v_p^i)$, $\nu_k$ are nodes sampled from $\bigcup\limits_{l\in \mathbb{N}^*} \mathcal{N}^l (v_q^j)$, $\lambda_i$ are the eigenvalues of $\mathbf{W}^T \mathbf{W}$, and $\mathbf{W} \in \mathbb{R}^{(l_r+1)d_v \times (l_a d_a + l_r d_r)}$ is constructed with the matrices in $\mathrm{AFE}_{\mathrm{node}}$ and $\mathrm{AFE}_{\mathrm{struct}}$. Specifically, $\mathbf{W}$ is a block matrix constructed as follows: 

1. $\mathbf{W}_{1:l_ad_a,1:d_v}$ are filled by the concatenation of $\{\mathbf{A}_i | i=1,...,l_a \}$, i.e. $[\mathbf{A}_1;...;\mathbf{A}_{l_a}] \in \mathbb{R}^{l_a d_a \times d_v}$.

2. For $\mathbf{W}_{l_ad_a+1:l_ad_a+l_rd_r, 1:(l_r+1)d_v}$, the construction is first filling $\mathbf{W}_{l_ad_a+(k-1)d_r:l_ad_a+kd_r, kd_v:(k+1)d_v}$ with $\mathbf{R}_k$, $k=1,...,l_r$.

3. For other parts, fill with zeros. 
\end{lemma}

\begin{proof}
Given vertex $v_p^i$, we concatenate its feature vector with the $l_r$ neighbors sampled from $\bigcup\limits_{l\in \mathbb{N}^*} \mathcal{N}^l (v_p^i)$, i.e. $\mathbf{x}_{p,i}' = [\mathbf{x}(v_p^i); \mathbf{x}(u_1); ...; \mathbf{x}(u_{l_r})]\in \mathbb{R}^{(l_r+1)d \times 1}, u_j \in \bigcup\limits_{l\in \mathbb{N}^*} \mathcal{N}^l (v_p^i)$. Then with the constructed block matrix $\mathbf{W}$, we could formulate the generation of $\mathbf{z}_p^i$ as: $\mathbf{z}_p^i = \mathbf{W} \mathbf{x}_{p,i}'$.

Similarly, we can formulate $\mathbf{z}_q^j$ for another vertex $v_q^j$.
And their distance can be formulated as:
\begin{align*}
\scriptsize
    \mathrm{dist}(\mathbf{z}_p^i,\mathbf{z}_q^j) = \norm{\mathbf{z}_p^i-\mathbf{z}_q^j}_2 = \sqrt{(\mathbf{z}_p^i-\mathbf{z}_q^j)^T(\mathbf{z}_p^i-\mathbf{z}_q^j)},
\end{align*}
where $\mathbf{z}_p^i-\mathbf{z}_q^j)^T(\mathbf{z}_p^i-\mathbf{z}_q^j)$ can be further expanded as:
\begin{flalign*}
    (\mathbf{z}_p^i-\mathbf{z}_q^j)^T(\mathbf{z}_p^i-\mathbf{z}_q^j) &
= (\mathbf{W} \mathbf{x}_{p,i}'-\mathbf{W} \mathbf{x}_{q,j}')^T (\mathbf{W} \mathbf{x}_{p,i}'-\mathbf{W} \mathbf{x}_{q,j}') &&\\
&=\big(\mathbf{W} (\mathbf{x}_{p,i}'- \mathbf{x}_{q,j}')\big)^T \big(\mathbf{W} (\mathbf{x}_{p,i}'- \mathbf{x}_{q,j}')\big)  &&\\
&= (\mathbf{x}_{p,i}'- \mathbf{x}_{q,j}')^T\mathbf{W}^T \mathbf{W} (\mathbf{x}_{p,i}'- \mathbf{x}_{q,j}')&& 
\end{flalign*}

According to lemma \ref{real_symmetric}, $\mathbf{W}^T \mathbf{W}$ is a real symmetric matrix, with lemma \ref{quadrtic_bound}, we have

\begin{align*}
    \frac{(\mathbf{x}_{p,i}'- \mathbf{x}_{q,j}')^T\mathbf{W}^T \mathbf{W} (\mathbf{x}_{p,i}'- \mathbf{x}_{q,j}')}{(\mathbf{x}_{p,i}'- \mathbf{x}_{q,j}')^T (\mathbf{x}_{p,i}'- \mathbf{x}_{q,j}')} \geqslant \lambda_{min}
\end{align*}\vspace{-2mm}

According to lemma \ref{real_symmetric}, with $l_ad_a+l_rd_r \geqslant (l_r+1)d_v $ and the constraint of column full rank on $\mathbf{W}$, $\mathbf{W}^T \mathbf{W} \in \mathbb{R}^{(l_r+1) \times (l_r+1)}$ has $l_r+1$ positive eigenvalues, thus $\lambda_{min} > 0$. 
Then we decompose $(\mathbf{x}_{p,i}'- \mathbf{x}_{q,j}')^T (\mathbf{x}_{p,i}'- \mathbf{x}_{q,j}')$ as:
\begin{flalign*}
    &\quad (\mathbf{x}_{p,i}'- \mathbf{x}_{q,j}')^T (\mathbf{x}_{p,i}'- \mathbf{x}_{q,j}')&& \\
    &= \sum_{k=0}^{(l_r+1)d-1} \big((\mathbf{x}_{p,i}')_k - (\mathbf{x}_{q,j}')\big)^2&&\\
    &= \sum_{k=1}^{(l_r+1)d_v}\big((\mathbf{x}_{p,i}')_k - (\mathbf{x}_{q,j}')\big)^2 &&\\
    &= \sum_{k=1}^{d_v}\big((\mathbf{x}_{p,i}')_k - (\mathbf{x}_{q,j}')_k\big)^2 + \sum_{k=1}^{l_r}\sum_{m=kd_v+1}^{(k+1)d_v}\big((\mathbf{x}_{p,i}')_k - (\mathbf{x}_{q,j}')_k\big)^2 &&\\
    &= \norm{\mathbf{x}(v_p^i)-\mathbf{x}(v_q^j)}_2^2+ \sum_{m=1}^{l_r}\norm{\mathbf{x}(u_m)-\mathbf{x}(\nu_k)}_2^2&&
\end{flalign*}

\noindent$\therefore \mathrm{dist}^2(\mathbf{z}_p^i - \mathbf{z}_q^j) = \norm{\mathbf{z}_p^i - \mathbf{z}_q^j}_2^2 \geqslant \lambda_{\min}(\norm{\mathbf{x}(v_p^i)-\mathbf{x}(v_q^j)}_2^2 + \sum_{k=1}^{l_r}\norm{\mathbf{x}(u_k)-\mathbf{x}(\nu_k)}_2^2)$ \vspace{-1mm}

\end{proof}\vspace{-2mm}

The key point in these theories is that for any task sequence with certain distance among the tasks, there exists a configuration that ensures HPNs to be capable of preserving the task distance after projecting the data into the hidden space, so that only the prototypes associated with the current task are refined and the prototypes corresponding to the other tasks are preserved. Specifically, theorem on zero-forgetting can be formulated as follows:
\begin{thm}[Task distance preserving]\label{zero-forget}
For $\mathrm{HPNs}$ trained on consecutive tasks $\mathcal{T}^p$ and $\mathcal{T}^{p+1}$. If $l_ad_a+l_rd_r \geqslant (l_r+1)d_v $ and $\mathbf{W}$ is column full rank, then as long as
$ t_{A} < \lambda_{\min}(l_r+1) \mathbf{dist}(\mathbb{V}_p, \mathbb{V}_{p+1}) $,  learning on $\mathcal{T}^{p+1}$ will not modify representations $\mathrm{HPNs}$ generate for data from $\mathcal{T}^p$, i.e. catastrophic forgetting is avoided. 
\end{thm}
In Theorem \ref{zero-forget}, $\lambda_i$ is eigenvalues of the $\mathbf{W}^T \mathbf{W}$, where $\mathbf{W}$ is the matrix mentioned before constructed via AFEs. $d_v$, $d_a$ and $d_r$ are dimensions of data, atomic node embeddings, and atomic structure embeddings. 
\begin{proof}
Following the proofs above, suppose two nodes $v_p^i$ and $v_q^j$ are embedded into $\mathbf{z}_p^i$ and $\mathbf{z}_q^j$ with the embedding module. Then the distance between $\mathbf{z}_p^i$ and $\mathbf{z}_q^j$ could be formulated as:

$\mathrm{dist}(\mathbf{z}_p^i,\mathbf{z}_q^j) = ||\mathbf{z}_p^i-\mathbf{z}_q^j||_2 = \sqrt{(\mathbf{z}_p^i-\mathbf{z}_q^j)^T(\mathbf{z}_p^i-\mathbf{z}_q^j)}$

According to lemma \ref{dist_bound}, we have $\mathrm{dist}^2(\mathbf{z}_p^i, \mathbf{z}_q^j) = ||\mathbf{z}_p^i - \mathbf{z}_q^j||_2^2 \geqslant \lambda_{\min}(||\mathbf{x}(v_p^i)-\mathbf{x}(v_q^j)||_2^2 + \sum_{k=1}^{l_r}||\mathbf{x}(u_k)-\mathbf{x}(\nu_k)||_2^2)$.

\noindent$\because v_p^i \in \mathbb{V}_p, v_q^j \in \mathbb{V}_q$, 

\noindent$\therefore ||\mathbf{x}(v_p^i)-\mathbf{x}(v_q^j)||_2^2 \geqslant \mathrm{dist}^2(\mathbb{V}_p, \mathbb{V}_q)$.

Similarly, $||\mathbf{x}(u_k)-\mathbf{x}(\nu_k)||_2^2 \geqslant \mathrm{dist}^2(\mathbb{V}_p, \mathbb{V}_q)$, for $\forall k$. 

\noindent$\therefore ||\mathbf{z}_p^i, \mathbf{z}_q^j||_2^2 \geqslant \lambda_{\min}(l_r+1) \mathrm{dist}^2(\mathbb{V}_p, \mathbb{V}_q)$ 

\noindent$\therefore \mathrm{dist}(\mathbf{z}_p^i, \mathbf{z}_q^j) =||\mathbf{z}_p^i - \mathbf{z}_q^j||_2 \geqslant \sqrt{\lambda_{\min}(l_r+1) \mathrm{dist}^2(\mathbb{V}_p, \mathbb{V}_q)} $

\noindent$\therefore$ If $t_A < \sqrt{\lambda_{\min}(l_r+1) \mathrm{dist}^2(\mathbb{V}_p, \mathbb{V}_q)}$, the embeddings of two nodes from two different tasks will not be assigned to same A-prototypes.

Above all, if the conditions in Theorem \ref{zero-forget} are satisfied, learning on new tasks will not modify the prototypes for previous tasks.
Besides, the data from previous tasks will be exactly matched to the correct prototypes after training the model on new tasks. In practice, the conditions may not be easy to be satisfied all the time. However, as mentioned in the paper, the bound given in Theorem \ref{zero-forget} is not tight, thus fully satisfying the conditions may not be necessary. Therefore, in the experimental section in the paper, we practically show how the important factors included in these conditions influence the performance (Section 3.6 in the paper). The results demonstrates that the more we satisfy the conditions, the better performance we will obtain, and certain factors (number of AFEs) influence more than the others.
\end{proof}\vspace{-2mm}

\begin{rmrk}
When $\mathrm{dist}(\mathbb{V}_p, \mathbb{V}_q) = 0$, i.e. there exists a non-empty set $\mathbb{V}_{\cap} = \mathbb{V}_p \cap \mathbb{V}_q$, $\mathrm{st.}$ $\mathrm{dist}(\mathbb{V}_p \setminus \mathbb{V}_{\cap}, \mathbb{V}_q \setminus \mathbb{V}_{\cap}) > 0$, then \textrm{Theorem} \ref{zero-forget} holds. As for the $\mathbb{V}_{\cap}$ containing examples exactly same in $\mathbb{V}_p$ and $\mathbb{V}_q$, there are two situations: 

1. $\forall v \in \mathbb{V}_{\cap}$, $y_p(v) = y_q(v)$, where $y_p(\cdot)$ and $y_q(\cdot)$ denote the associated labels in task $p$ and $q$

2. $\exists v \in \mathbb{V}_{\cap}$, $y_p(v) \neq y_q(v)$

For situation 1, $\mathbb{V}_{\cap}$ will not cause forgetting on the previous task, as these shared data are exactly same and will optimize the model to same direction. 
For situation 2, if no task indicator is provided, then these data are contradictory data, if task indicator is provided, then the indicator could be merged into the feature vector of the node, i.e. $\mathbf{x}(v_p)$, then $v_p$ will not belong to $\mathbb{V}_{\cap}$.\vspace{-2mm}
\end{rmrk}


%

\ifCLASSOPTIONcaptionsoff
  \newpage
\fi



\bibliographystyle{IEEEtran}
\bibliography{ref}
%

%

\end{document}


\title{Supplementary Materials of Hierarchical Prototype Networks for Continual Graph Representation Learning}

\maketitle
\IEEEPARstart{I}{n} this document, we provide implementation details in Section \ref{sec:implementation_details} and additional experimental results and analysis in Section \ref{sec:experiments}.

\section{Details of Implementation}\label{sec:implementation_details}

\subsection{Datasets and task splitting}
In this subsection, we introduce the datasets we used and the details of how each dataset is split into different tasks.

We use 5 publicly datasets which include 2 citation networks (Cora\cite{sen2008collective}, Citeseer \cite{sen2008collective}, OGB-Arxiv \cite{wang2020microsoft,mikolov2013distributed}), 1 actor co-occurrence network (Actor) \cite{pei2020geom}, and 1 product co-purchasing network (OGB-Products \cite{Bhatia16}). 

\subsubsection{Citation networks}
The original Cora \cite{mccallum2000automating} and Citeseer \cite{giles1998citeseer} are pre-processed by Sen et al. \cite{sen2008collective} with stemming and removing stop words as well as words with document frequency less than 10. Finally, Cora contains 2708 documents, 5429 links denoting the citations among the documents, and each document is represented with 1433 distinct words. Cora contains 7 classes. For training, 140 documents are selected with 20 examples for each class. The validation set contains 500 documents and the test set contains 1000 examples. In our continual learning setting, the first 6 classes are selected and grouped into 3 tasks (2 classes for each task) in the original order.
Citeseer results in 3312 documents with each document being represented with 3703 distinct words, and 4732 links. Citeseer contains 6 classes. 20 documents per class are selected, for training, 500 documents are selected, for validation, and 1000 documents are selected as the test set. For continual learning setting, the documents from 6 classes are grouped into 3 tasks with 2 classes per task in the original order.
The Cora and Citeseer datasets can be downloaded via \href{https://github.com/tkipf/gcn/tree/master/gcn/data}{Cora$\&$Citeseer}.

The OGB-Arxiv dataset is collected in the Open Graph Benchmark \href{https://ogb.stanford.edu/docs/nodeprop/#ogbn-arxiv}{OGB}. It is a directed citation network between all Computer Science (CS) arXiv papers indexed by MAG \cite{wang2020microsoft}. Totally it contains 169,343 nodes and 1,166,243 edges. Each node is an paper and each directed edge indicates that one paper cites another one. Each paper comes with a 128-dimensional feature vector. The dataset contains 40 classes. As the dataset is not balanced and the numbers of examples in different classes differs significantly, directly grouping the classes into 2-class groups like the Cora and Citeseer will cause certain tasks to be imbalanced. Therefore, we reordered the classes in an descending order according to the number of examples contained in each class, and then group the classes according to the new order. In this way, the number of examples contained in different classes of each task are arranged to be as balanced as possible. Specifically, the class indices of each task are: \{(35, 12),(15, 21),(28, 30), (16, 24), (10, 34), (8, 4), (5, 2), (27, 26), (36, 19), (23, 31), (9, 37), (13, 3), (20, 39), (22, 6), (38, 33), (25, 11), (18, 1), (14, 7), (0, 17), (29, 32)\}.

\subsubsection{Actor co-occurrence network}
The actor co-occurrence network is a subgraph of the film-director-actor-writer network \cite{tang2009social}. Each node in this dataset corresponds to an author, and the edges between the nodes are co-occurrence on the same Wikipedia pages. The whole dataset contains 7600 nodes and 33544 edges. Each node is accompanied with a feature vector of 931 dimensions. The nodes are classified into 4 classes according to the number of the average monthly traffic of the web page. For this dataset, we also constructed 2 tasks with 2 classes per task. The link to this dataset is \href{https://github.com/graphdml-uiuc-jlu/geom-gcn/tree/master/new_data/film}{Actor}. The balanced splitting of the classes is \{(0, 1), (2, 3)\}.

\subsubsection{Product co-purchasing network}
OGB-Products is also collected in the Open Graph Benchmark \href{https://ogb.stanford.edu/docs/nodeprop/#ogbn-arxiv}{OGB}, and is an undirected and unweighted graph, representing an Amazon product co-purchasing network \href{http://manikvarma.org/downloads/XC/XMLRepository.html}{link}. In total, it contains 2,449,029 nodes and 61,859,140 edges. Nodes represent products sold in Amazon, and edges between two products indicate that the products are purchased together. Node features are generated by extracting bag-of-words features from the product descriptions followed by a Principal Component Analysis to reduce the dimension to 100. 47 top-level categories are used for target labels, in our experiments, we select 46 classes and omit the final class containing only 1 example. Similar to OGB-Arxiv, we reorder the classes in an descending order according to the number of examples contained in each class, and then group the classes according to the new order. The class indices of each tasks are: \{(4, 7), (6, 3), (12, 2), (0, 8), (1, 13), (16, 21), (9, 10), (18, 24), (17, 5), (11, 42), (15, 20), (19, 23), (14, 25), (28, 29), (43, 22), (36, 44), (26, 37), (32, 31), (30, 27), (34, 38), (41, 35), (39, 33), (45, 40)\}.

\begin{figure*}
    \centering
    \includegraphics[height=8cm]{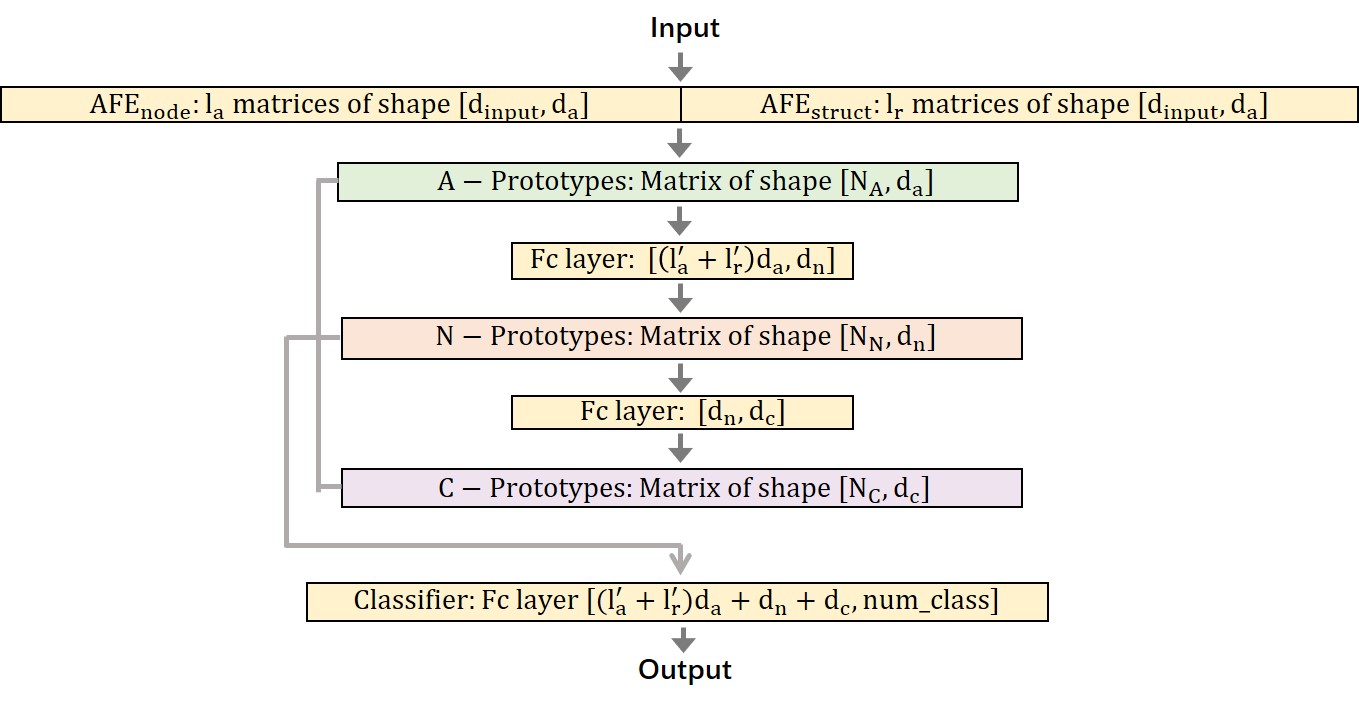}
    \caption{Details of modules in HPNs.}
    \label{fig:module_structure}
\end{figure*}

\subsection{Experiment Setup}
All models are implemented in PyTorch with SGD optimizer and repeated 5 times on a Nvidia Titan Xp GPU. The average performance and standard deviations are reported for comparison. The network architecture of HPNs is detailed in Figure \ref{fig:module_structure}, and the specific values of hyperparameters will be given in the following. The hyperparameters we provide here correspond to the models used in comparisons with the baselines, while in other experiments the hyperparameters are the research objects and will not be kept unchanged.

As the sizes of the datasets we used are greatly different, we adopt different hyperparameters for small datasets and large datasets. 
The small datasets include Cora, Citeseer, Actor, Wisconsin, Cornell, and Texas. The large datasets include OGB-Arxiv and OGB-Products. 

For the small datasets, we set $l_a'=1$, $l_r'=1$, and $h=2$. We randomly sample 5 one-hop neighbors and 7 two-hop neighbors. The learning rates are managed separately for different modules of the model. For the AFEs, the learning rate is set as 0.1 at the beginning and decays to 0.001 at epoch 35. The learning rate for the prototypes are initialized as 0.1 at epoch 35 and decays to 0.01 at epoch 85. And the learning rates for the other trainable parameters are the same as the AFEs. During training, the AFEs would change rapidly at first and slow down after several epochs. Therefore, at the starting period of training, the same node would not be stably matched to the same set of prototypes due to the rapidly changing AFEs. To avoid this from creating too many redundant prototypes, we start to establish prototypes after training the AFEs at the 35th epochs. The input data has a dimension of 1433, and we set the dimensions of A-, N-, and C-prototypes to be 16. The number of training epochs is 90. Although the training procedure is designed in a delicate way, the model is actually rather robust and can perform well without these delicate procedures. For example, on the largest dataset OGB-Products, we only train the model for 10 epochs, and do not decay the learning rate, the prototypes are established at the beginning, and the model still obtained good results, as shown in in the paper (e.g. results in Section 3.3). For the large datasets, for higher efficiency, we set $h=1$, and only uniformly sample one neighbor from the neighbors. On the OGB-Products, we shrink the dimensions of A-, N-, and C-prototypes to be 2, in order to control the number of prototypes.
For both HPNs, the threshold $t_A$, $t_N$, and $t_C$ are selected by cross validation on $\{0.01, 0.05,0.1,0.15,0.2,0.25,0.3,0.35, 0.4\}$. According to the experimental results, there is a wide range for choosing the thresholds. Finally we choose $t_A=t_N=0.3$ and $t_C=0.4$. 

Figure \ref{fig:module_structure} describes the shapes of the modules in HPNs. The input data are $\mathrm{d_{input}}$ node feature vectors. With the $\mathrm{AFEs}$, the data are transformed into atomic embeddings, which are $\mathrm{d_a}$ dimensional vectors. Then they are mapped to A-prototypes with same dimensions. After that, the matched A-prototypes are further mapped to higher level N- and C-prototypes with the corresponding Fc layers. The dimensions of N- and C- prototypes are $\mathrm{d_n}$ and $\mathrm{d_c}$. Finally, all the prototypes of different levels are concatenated into a single vector with length of $\mathrm{(l_a'+l_r')d_a+d_n+d_c}$ and fed into the classifier (Fc layer) for classification results. The number of logits output by the classifier is $\mathrm{num\_{class}}$, which is the number of classes in each task. Some existing continual learning works with the task-incremental setting will expand the number of logits output by the classifier to $\mathrm{num\_{class}}\cdot \mathrm{num\_{task}}$, where $\mathrm{num\_task}$ is the total number of tasks the model is going to encounter. however, we argue that this is a very impractical scenario because of the following reasons: 1. A continual learning model should not know the number of tasks to learn in advance, therefore $\mathrm{num\_{task}}$ is unknown. 2. Setting the number of logits as $\mathrm{num\_{class}}\cdot \mathrm{num\_{task}}$ causes the memory consumption of the model to grows linearly with the number of tasks to learn, which is highly undesirable for continual learning models. Therefore, we set the number of output logits as $\mathrm{num\_{class}}$ and force different tasks to share the output head, which increases the hardness of learning but is much more practical.

The baselines have different settings. For the baselines with GCN backbone, 16 is approximately the best for the number of hidden unit. For the GAT based baselines, we set the number of heads and number of hidden units as 8. For GIN, the number of hidden units is 32. For all these baselines, the above mentioned settings are applied on most datasets. For some datasets on which the baselines cannot perform well, we will further tune the models carefully to get better results.

\section{Additional Experimental Results and Detailed Analysis}\label{sec:experiments}
To further validate our proposed model, in this section, we report additional experimental results by extending the experiments reported in the paper to more datasets. We will also give more detailed analysis on the results, which is omitted in the paper due to space limitations.

\begin{figure}[h]
    \centering
        \centering
        \includegraphics[height=4cm]{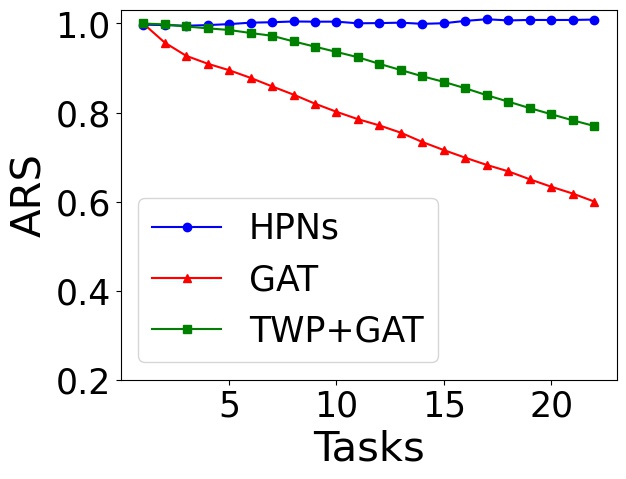}
        \caption{Dynamics of ARS for continual learning tasks on OGB-Products dataset.}
        \label{fig:ARS_Memory}
\end{figure}
\begin{figure}[h]
        \centering
        \includegraphics[height=4cm]{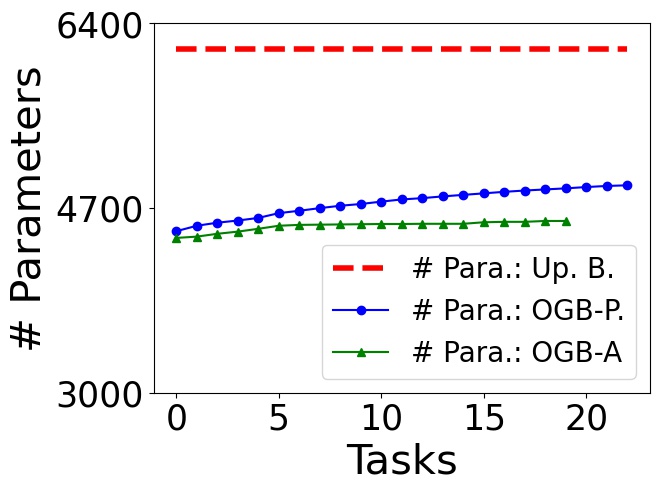}
     \caption{Dynamics of memory consumption of HPNs on both OGB-Arxiv and OGB-Products.}
     \label{fig:Param_amount}
\end{figure}

\begin{figure*}[h]
    \centering
    \begin{minipage}{0.33\textwidth}
        \centering
        \includegraphics[width=1.\textwidth]{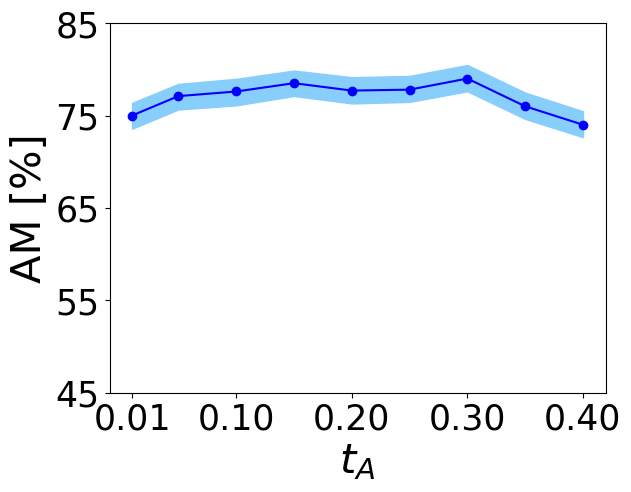}
    \end{minipage}\hfill
    \begin{minipage}{0.33\textwidth}
        \centering
        \includegraphics[width=1.\textwidth]{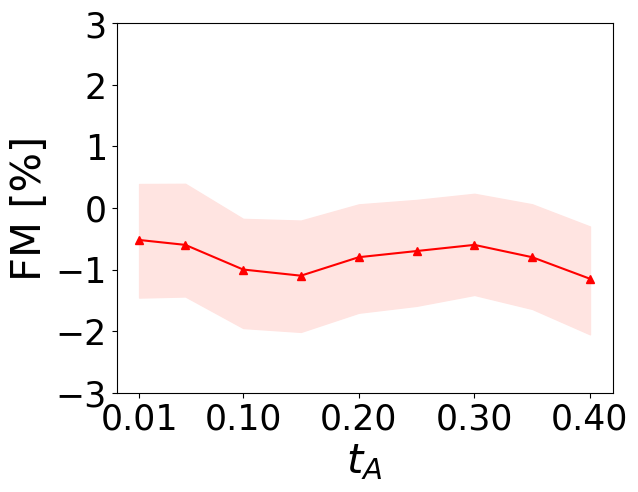}
    \end{minipage}\hfill
    \begin{minipage}{0.33\textwidth}
        \centering
        \includegraphics[width=1.\textwidth]{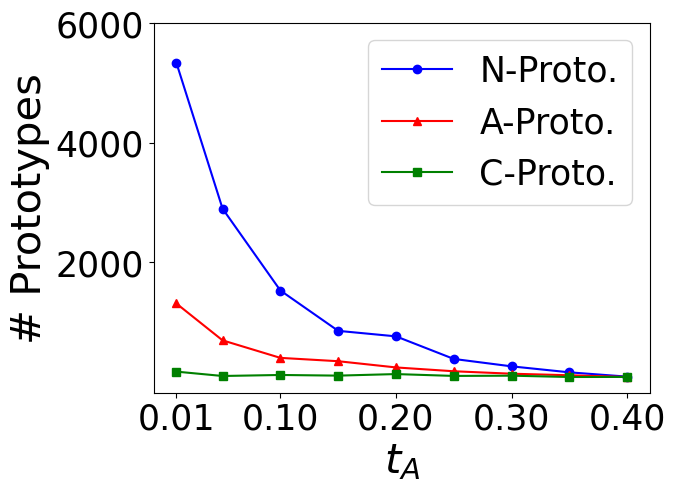}
    \end{minipage}
     \caption{Left and Middle: AM and FM change when $t_A$ varies on Citeseer. Right: impact of $t_A$ on the number of prototypes over Citeseer.}
     \label{fig:parameter_sensitivity}
\end{figure*}

\begin{figure*}[h]
    \centering
    \begin{minipage}{0.29\textwidth}
        \centering
        \includegraphics[width=1.\textwidth]{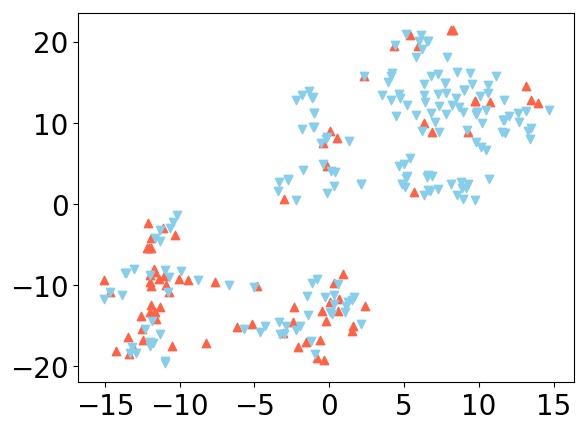}
    \end{minipage}\hfill
    \begin{minipage}{0.29\textwidth}
        \centering
        \includegraphics[width=1.\textwidth]{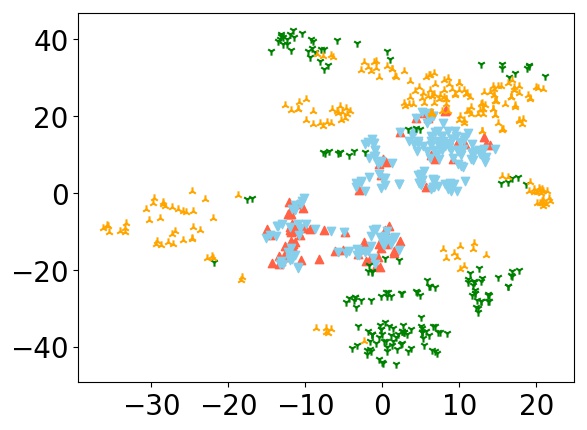}
    \end{minipage}\hfill
    \begin{minipage}{0.4\textwidth}
        \centering
        \includegraphics[width=1.\textwidth]{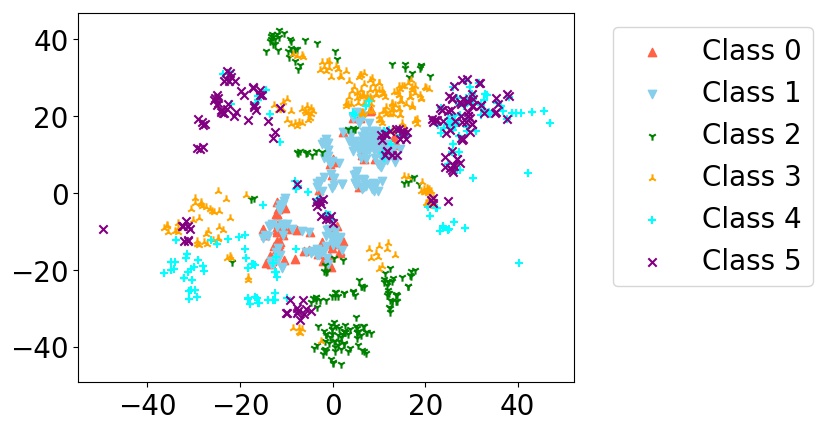}
    \end{minipage}
     \caption{Visualization of hierarchical prototype representations of test data of different tasks from Citeseer via t-SNE.}
     \label{fig:tsne_citeseer}
\end{figure*}

\subsection{Additional Results on Ablation Study}
In this subsection, we provide the ablation study results on another large dataset OGB-Arxiv, and the results are shown in Table \ref{tab:ablation_proto} and \ref{tab:ablation_loss}. 
\begin{table}[]
    \centering
    
    \caption{Ablation study on prototypes of different levels of prototypes over OGB-Arxiv.}
    \begin{tabular}{c|c|c|c|c|c}
    \toprule
     Conf. &A-p.     & N-p.    & C-p.    & AM\%           & FM\% \\ \midrule
     1  &\checkmark &           &           &  82.1$\pm$0.9    &  +0.0$\pm$1.1 \\ \midrule
     2  &\checkmark &\checkmark &           &  83.6$\pm$1.2    &  +0.2$\pm$0.9   \\ \midrule
     3  &\checkmark &\checkmark &\checkmark &  85.8$\pm$0.7    &  +0.6$\pm$0.9 \\ \bottomrule
    \end{tabular}
    \label{tab:ablation_proto}
\end{table}

\begin{table}[]
    \centering
    \caption{Ablation study on different loss terms over OGB-Arxiv.}
    \begin{tabular}{c|c|c|c|c|c}
    \toprule
     Conf.  &$\mathcal{L}_{cls}$ & $\mathcal{L}_{div}$ & $\mathcal{L}_{dis}$ & AM\%        & FM\%        \\ \midrule
     1 &\checkmark           &                     &                     &79.6$\pm$1.5 &-0.3$\pm$1.3  \\ \midrule
     2 &\checkmark           & \checkmark          &                     &82.3$\pm$1.0 &+0.4$\pm$0.9 \\ \midrule
     3 &\checkmark           &                     &\checkmark           &80.7$\pm$1.2 &+0.0$\pm$1.4  \\ \midrule
     4 &\checkmark           &\checkmark           &\checkmark           &  85.8$\pm$0.7    &  +0.6$\pm$0.9   \\ \bottomrule
    \end{tabular}
    \label{tab:ablation_loss}
\end{table}

From Table \ref{tab:ablation_proto}, we can observe that on the large dataset, the improvements brought by high level prototypes are more significant than on the small dataset (reported in Table 2 in the paper). Similarly, in Table \ref{tab:ablation_loss}, the influence of different loss terms is also more prominent compared to the results reported in Table 3 in the paper. The above results imply that our proposed hierarchical prototypes and different loss terms are effective, and the effectiveness becomes increasingly significant on larger datasets with richer information.

\subsection{Additional Results on Learning Dynamics}
Besides the learning dynamics on OGB-Arxiv provided in Section 3.5 in the paper, we further provide the results on OGB-Products, as shown in Figure \ref{fig:ARS_Memory}.

The learning dynamics shown in Figure \ref{fig:Param_amount} is similar to the one on OGB-Products shown in the paper. The only difference is that OGB-Products contains more tasks and the ARS of the baselines decrease more than on OGB-Arxiv.

\subsection{Additional Results on Parameter Sensitivity}
In Figure \ref{fig:parameter_sensitivity}, we further provide the parameter sensitivity results on citeseer dataset. The results have similar patterns with the results provided in the paper.

\subsection{Additional Results on Memory Consumption}

In Figure \ref{fig:Param_amount}, we simultaneously show the memory consumption change via the number of tasks on both OGB-Arxiv and OGB-Products. We use same model configurations for both datasets, thus the upper bounds of the memory consumption are same. From Figure \ref{fig:ARS_Memory}, we could see that the memory consumption of HPNs on both datasets increases slowly and far less than the upper bound. Although OGB-Products is more than ten times larger than OGB-Arxiv, the memory used on OGB-Products is only slightly more than on OGB-Arxiv. demonstrating the memory efficiency of HPNs.

\subsection{Additional Results on Visualization}

In Figure \ref{fig:tsne_citeseer}, we visualize the hierarchical prototype representations of the test nodes on Citeseer by t-SNE~\cite{van2008visualizing}. Similar to the visualization results shown in the paper, Figure \ref{fig:tsne_citeseer} sequentially show the classes of task 1 (Left), task 1,2 (Middle), and task 1,2,3 (Right). The examples belonging to different classes are denoted with different shapes and colors, as shown in the legend on the right.

\ifCLASSOPTIONcaptionsoff
  \newpage
\fi

\bibliographystyle{IEEEtran}
\bibliography{supplementary}